\documentclass{article}

\usepackage[position, final]{neurips_2026}

\usepackage[utf8]{inputenc}
\usepackage[T1]{fontenc}
\usepackage{hyperref}
\usepackage{url}
\usepackage{booktabs}
\usepackage{amsfonts}
\usepackage{amsmath}
\usepackage{nicefrac}
\usepackage{microtype}
\usepackage{xcolor}
\usepackage{graphicx}
\usepackage{tikz}
\usetikzlibrary{arrows.meta, positioning, shapes.geometric, fit, backgrounds}
\usepackage{pgfplots}
\usepackage{pgfplotstable}
\pgfplotsset{compat=1.18,
  every axis/.append style={
    /pgf/number format/1000 sep={},
    /pgf/number format/fixed,
  }
}
\usepgfplotslibrary{groupplots}

\title{Machine Learning Research Has Outpaced Its Communication Norms and NeurIPS Should Act}

\author{%
  Ajay Manydam Rangarajan \\
  Independent Researcher \\
  \texttt{ajay.rangarajan@rwth-aachen.de} \\
  \And
  Jeyashree Krishnan \\
  Institute for Computational Biomedicine,\\ University Hospital Aachen \\
  \texttt{jeyashree.krishnan@rwth-aachen.de}
}

\begin{document}
\maketitle

% Abstract.
% Position paper rule: abstract must briefly state the position using the
% formula ``This position paper argues that <statement>.''
\begin{abstract}
  \textbf{This position paper argues that machine learning research
  has grown exponentially while its communication norms have not,
  and that NeurIPS could adopt explicit, measurable writing
  standards of the kind we propose in this paper.}
  We analyze the titles and abstracts of roughly 2.8 million arXiv
  papers (1991--2025), 24,772 NeurIPS papers
  (1987--2024), and 24.5 million PubMed papers
  (1990--2025).
  We apply five families of metrics: classical readability scores,
  the Hohmann writing style suite (which includes sensational
  language), acronym density and reuse, an LLM as judge readability
  protocol, and citation counts joined from OpenAlex and Semantic
  Scholar.
  Four patterns emerge from the data that suggest research
  communication has become harder over time.
  First, NeurIPS abstracts score harder to read on every classical
  readability metric we test.
  Flesch Reading Ease falls from approximately 24 in 1987 to
  approximately 13 in 2024, and sensational language rates rise by
  approximately 50 percent in NeurIPS abstracts between 2015 and
  2024.
  Second, acronym density in NeurIPS titles and abstracts has grown
  substantially, with title density rising from 0.33 per 100 words
  in 1987 to 3.21 in 2024, and roughly 89 percent of NeurIPS
  acronyms used fewer than ten times, ten percentage points above
  the science-wide baseline.
  Third, more readable NeurIPS papers tend to receive more
  citations, suggesting that readability and research impact are
  correlated, and that less readable papers risk remaining
  fragmented within the literature.
  LLM as judge scores rate NeurIPS abstracts as approximately
  stable from 1987 to roughly 2022, with early signs of
  improvement thereafter, a pattern that disagrees with every
  classical readability metric.
  This disagreement raises a design question for any enforcement
  policy of whether the target reader is a human or an LLM.
  Lastly, NeurIPS paper volume has grown by a factor of roughly 50
  between 1987 and 2024, from fewer than 100 accepted papers to
  approximately 4,500.
  Assuming the goal is to optimise for human readers, we propose
  seven standards NeurIPS could pilot at NeurIPS~2027: an acronym
  budget paired with a venue-maintained approved-term list, a
  human readability threshold at submission time, stricter
  citation standards, a requirement for standalone visual
  elements, a plain language summary, a pre-registered acronym
  glossary covering the full paper, and open source audit
  tooling.
\end{abstract}

% 1. Introduction.
\section{Introduction}
\label{sec:intro}

\textbf{Machine learning research has outgrown its communication
norms. NeurIPS, as the flagship venue of the field, should
enforce measurable standards on acronym introduction and reuse,
readability of titles and abstracts, citation quality and count,
and acceptance thresholds, and should back those standards with
open source tooling that makes every standard auditable by
authors, reviewers, and programme chairs at submission time.}

The output of machine learning research is doubling every two
to three years, and most of that output is written. Titles
become denser strings of acronyms, abstracts harder to parse,
and reviewers face dozens of submissions per cycle with
diminishing attention per paper. \citet{jeschke2019knowledge}
call the general version of this pattern a knowledge ignorance
paradox in which information grows but actionable understanding
does not. NeurIPS is interdisciplinary by design, serving theorists,
practitioners, clinical researchers, and policy analysts, and
the cost of opaque writing compounds across that audience.
Medical publishing confronted analogous pressures decades ago
and responded with explicit editorial style guides
\citep{berlin2013acronym}; machine learning has not yet adopted
comparable venue-level norms.

We test this empirically across a corpus of roughly 2.8 million
arXiv papers (1991--2025), 24,772 NeurIPS papers (1987--2024),
and 24.5 million PubMed abstracts (1990--2025). arXiv papers
are classified as ML if their primary category is in
\texttt{cs.LG}, \texttt{cs.AI}, \texttt{cs.CV}, \texttt{cs.CL},
\texttt{cs.NE}, \texttt{cs.RO}, \texttt{cs.IR}, \texttt{cs.MA},
\texttt{cs.HC}, \texttt{cs.CY}, or \texttt{stat.ML}. All
remaining arXiv categories form the non-ML control group, so
ML-specific trends can be separated from field-wide trends
without confounding venue and domain.

We investigate six hypotheses. First, that ML abstracts have
become harder to read on classical readability metrics. Second,
that LLM judges rate ML abstracts differently from classical
formulas, with implications for any automated enforcement
mechanism. Third, that acronym density has grown beyond the
science-wide baseline and most ML acronyms are used too
infrequently to become shared vocabulary. Fourth, that abstract
readability and citation count are positively correlated within
NeurIPS. Fifth, that sensational language rates have risen in
ML abstracts. Sixth, that publication volume has grown so
rapidly that existing peer review cannot absorb it without
changes to writing standards.

We apply five families of metrics. The first is 15 classical
readability scores \citep{plaven2017readability}. The second is
the nine-metric Hohmann writing style suite
\citep{hohmann2025evolution,hohmann2025comparing}. The third is
the acronym extraction algorithm of \citet{barnett2020growth}.
The fourth is the LLM as judge protocol of
\citet{cachola2025evaluating}. The fifth is citation counts for
NeurIPS papers from OpenAlex and Semantic Scholar. Each
measurable maps to one of four reader costs. Long sentences
impose working memory load \citep{britton1982sentence,
sweller1994cognitive}. Dense noun phrases force readers to hold
more concepts in parallel \citep{pinker2014sense}. Novel
acronyms occupy a memory slot until expansion is committed
\citep{shulman2020jargon}. Citation-dense sentences require
traversal of the reference graph. The full pipeline is
documented in Appendix~\ref{app:workflow}.

Two pressures sharpen the urgency at NeurIPS 2026. NeurIPS
volume has grown by a factor of roughly 50 since 1987, from
fewer than 100 accepted papers to approximately 4,500.
Widespread AI-assisted writing has arrived without auditing
mechanisms, and our data are consistent with a post-2022
inflection in jargon and sensational language that tracks the
availability of instruction-tuned writing assistants.
\citet{martinez2021specialized} show that specialised
terminology already reduces per-paper citation counts in
biology, suggesting the same dynamic now applies to ML.

\begin{figure}[h]
  \centering
  \resizebox{\linewidth}{!}{%
  \begin{tikzpicture}[
      node distance=7mm and 9mm,
      font=\scriptsize,
      every node/.style={align=center},
      source/.style={draw, rounded corners=1pt, fill=blue!8,
                     minimum width=18mm, minimum height=5mm, inner sep=1pt},
      stage/.style={draw, thick, rounded corners=2pt, fill=gray!10,
                    minimum width=26mm, minimum height=10mm, inner sep=3pt},
      metric/.style={draw, rounded corners=1pt, fill=orange!12,
                     minimum width=26mm, minimum height=5mm, inner sep=1pt},
      output/.style={draw, thick, rounded corners=2pt, fill=green!10,
                     minimum width=30mm, minimum height=10mm, inner sep=3pt},
      flow/.style={-{Latex[length=2mm]}, thick},
    ]
    % Column 1: sources.
    \node[source] (s1) {NeurIPS 1987--2024};
    \node[source, below=of s1] (s2) {arXiv 1991--2025 \\ {\footnotesize (ML and non-ML)}};
    \node[source, below=of s2] (s3) {PubMed 1990--2025};
    \node[above=1mm of s1, font=\scriptsize\bfseries] {Data sources};

    % Column 2: canonical schema.
    \node[stage, right=12mm of s2.east] (schema)
      {Canonical schema \\ (Pydantic validated) \\
       \texttt{paper\_id, venue, year,} \\
       \texttt{title, abstract, authors,} \\
       \texttt{arxiv\_primary\_category}};

    % Column 3: per paper metrics.
    \node[metric, right=14mm of schema.north east, yshift=-2mm] (m1)
      {Classical readability \\ (15 scores, \texttt{textstat})};
    \node[metric, below=2mm of m1] (m2)
      {Hohmann writing style \\ (9 metrics, spaCy)};
    \node[metric, below=2mm of m2] (m3)
      {Acronym density + reuse \\ (Barnett 2020 algorithm)};
    \node[metric, below=2mm of m3] (m4)
      {LLM as judge readability \\ (Cachola 2025 prompts)};
    \node[metric, below=2mm of m4] (m5)
      {Citations (OpenAlex, S2)};
    \node[above=1mm of m1, font=\scriptsize\bfseries] {Per paper metrics};

    % Column 4: aggregation.
    \node[stage, right=12mm of m3.east] (agg)
      {Aggregation \\ per venue-year \\ per arXiv category};

    % Column 5: outputs.
    \node[output, right=11mm of agg.north east, yshift=-5mm] (ev)
      {\textbf{Evidence} \\ volume, readability, \\ acronyms, fragmentation};
    \node[output, below=3mm of ev] (pol)
      {\textbf{Policy proposals} \\ seven NeurIPS standards};

    % Flow arrows: sources to schema.
    \foreach \src in {s1, s2, s3}
      \draw[flow] (\src.east) -- (schema.west);

    % Schema to metrics.
    \foreach \met in {m1, m2, m3, m4, m5}
      \draw[flow] (schema.east) -- (\met.west);

    % Metrics to aggregation.
    \foreach \met in {m1, m2, m3, m4, m5}
      \draw[flow] (\met.east) -- (agg.west);

    % Aggregation to outputs.
    \draw[flow] (agg.east) -- (ev.west);
    \draw[flow] (agg.east) -- (pol.west);
  \end{tikzpicture}%
  }
  \caption{\textbf{End to end workflow.} Three data sources feed a
  canonical Pydantic-validated schema. Five per-paper metric
  families are computed on titles and abstracts separately, then
  aggregated per venue-year and per arXiv primary category.
  The outputs feed the empirical evidence
  (Sections~\ref{sec:readability} through~\ref{sec:volume}) and
  the proposed policy standards
  (Section~\ref{sec:policy}). Pipeline details appear in
  Appendix~\ref{app:workflow}.}
  \label{fig:workflow}
\end{figure}
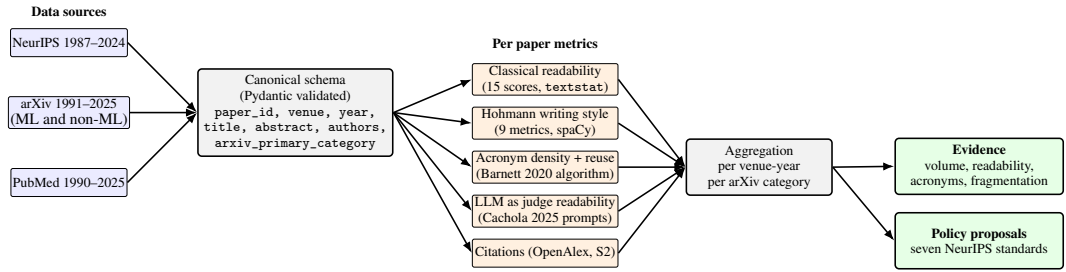

% 3. Readability.
\section{Readability: Papers Are Getting Harder to Read}
\label{sec:readability}

Readability of scientific abstracts is measurable through
established linguistic and cognitive formulas validated across
scientific domains \citep{plaven2017readability}. We apply three complementary families of
metrics: classical readability scores
(Section~\ref{sec:readability-classical}), sensational language
rates (Section~\ref{sec:sensational}), and the Hohmann writing
style suite (Section~\ref{sec:hohmann}). The classical
readability scores and sensational language rates show that ML
abstracts have become harder to read and more rhetorically loaded
over time, with the effect more pronounced in ML than in
comparable non-ML corpora. The Hohmann writing style suite is
mixed in direction. Several Hohmann metrics in fact move in the
direction of better writing, consistent with the wider scientific
publishing trend toward more active narration and clearer
signposting reported by
\citet{hohmann2025evolution,hohmann2025comparing}. The two
families together describe an ML literature in which surface
readability has fallen even as a few targeted style features have
improved.

\subsection{Classical readability scores}
\label{sec:readability-classical}

The Flesch Reading Ease score assigns values from 0 to 100,
where higher values indicate easier reading and lower values
indicate greater cognitive burden
\citep{plaven2017readability,sweller1994cognitive}.
A score near 10 corresponds to text requiring doctoral-level
reading proficiency. A score near 25 is considered very
difficult, appropriate for professional and academic audiences.
This metric is chosen as the primary indicator here because it
is the only one of the 15 we compute for which higher is better;
its decline is therefore unambiguous in direction.

\begin{figure}[h]
\centering
\begin{tikzpicture}
\begin{axis}[
    width=0.88\linewidth, height=5cm,
    xlabel={Year},
    ylabel={Flesch Reading Ease (\textuparrow\ easier)},
    xtick distance=4,
    x tick label style={rotate=45, anchor=east, font=\tiny,
                        /pgf/number format/1000 sep={}},
    ytick align=inside,
    legend columns=4,
    legend style={draw=none, fill=none, font=\scriptsize,
                  cells={anchor=west}, column sep=8pt,
                  at={(0.5,-0.28)}, anchor=north},
    tick label style={font=\tiny},
    label style={font=\small},
    enlarge x limits=0.02,
  ]
  \addplot[color={rgb,255:red,44;green,160;blue,44}, dashdotted, very thick]
    table[x=year, y=neurips, col sep=comma]
    {data/pgfplots/readability_flesch_ease.csv};
  \addlegendentry{NeurIPS}
  \addplot[color={rgb,255:red,148;green,103;blue,189}, solid, very thick]
    table[x=year, y=arxiv_ml, col sep=comma]
    {data/pgfplots/readability_flesch_ease.csv};
  \addlegendentry{arXiv ML}
  \addplot[color={rgb,255:red,140;green,86;blue,75}, dotted, thick]
    table[x=year, y=arxiv_nonml, col sep=comma]
    {data/pgfplots/readability_flesch_ease.csv};
  \addlegendentry{arXiv non-ML}
  \addplot[color={rgb,255:red,214;green,39;blue,40}, dashed, thick]
    table[x=year, y=pubmed, col sep=comma]
    {data/pgfplots/readability_flesch_ease.csv};
  \addlegendentry{PubMed}
  % ChatGPT launch marker (late 2022)
  \draw[black, dashed, semithick] (axis cs:2022,-5) -- (axis cs:2022,50);
  \node[font=\tiny, rotate=90, anchor=south west]
    at (axis cs:2022.2,18) {ChatGPT (2022)};
\end{axis}
\end{tikzpicture}
\caption{\textbf{Flesch Reading Ease on NeurIPS, arXiv ML, arXiv
  non-ML, and PubMed, 1987--2025.}
  Higher values indicate easier reading (\textuparrow).
  NeurIPS falls from approximately 25 in 1987 to approximately
  10 in 2024. arXiv ML follows a similar trajectory. The arXiv
  non-ML baseline declines more slowly, indicating that the
  accelerated decline is specific to ML venues. Vertical dashed
  line marks the late-2022 availability of widely-used
  instruction-tuned writing assistants.}
\label{fig:readability-flesch}
\end{figure}
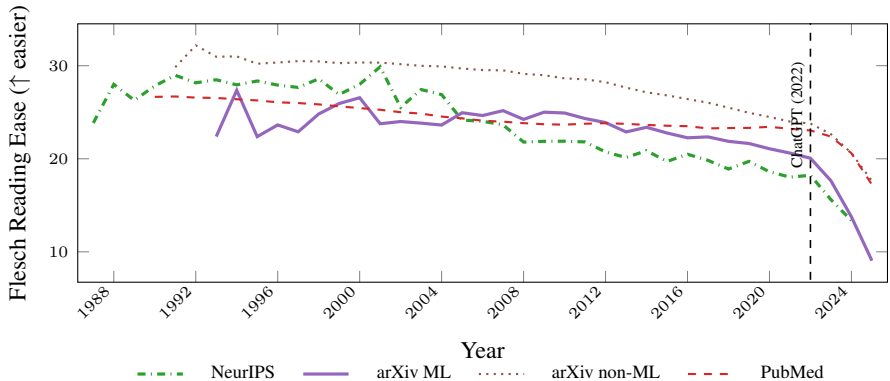

Figure~\ref{fig:readability-flesch} shows Flesch Reading Ease
on NeurIPS, arXiv ML, arXiv non-ML, and PubMed from 1987 to
2025. NeurIPS falls from approximately 24 in 1987 to
approximately 13 in 2024, a sustained decline across 37 years.
The arXiv ML aggregate follows a similar downward trajectory.
The arXiv non-ML baseline declines more slowly, confirming
that the accelerated decline is specific to ML rather than a
universal property of all scientific writing.

All 15 classical readability metrics we apply move in the same
direction on ML venues. Flesch-Kincaid Grade, Gunning Fog, SMOG,
Dale-Chall, Spache, Coleman-Liau, ARI, Linsear Write, LIX, RIX,
FORCAST, and Powers-Sumner-Kearl all rise on NeurIPS and arXiv ML
after approximately 2015 and accelerate after 2020. The vertical
marker in Figure~\ref{fig:readability-flesch} indicates the late
2022 availability of widely-used instruction-tuned writing
assistants. The rate of decline steepens after this point on ML
venues. The full figure with all 15 metrics appears in
Appendix~\ref{app:figures}.

% Flesch Reading Ease on NeurIPS, arXiv ML, arXiv non-ML, and PubMed.
% Data: data/pgfplots/readability_flesch_ease.csv
% Higher values = easier reading.

\subsection{Sensational language}
\label{sec:sensational}

\citet{millar2022sensational} define ten categories of
sensational language: importance, novelty, rigor, scale,
utility, quality, attitude, problem, additional emphasis, and
total count. \citet{hohmann2025evolution} incorporate these
categories as one component of their writing style suite. These
categories capture language that overstates certainty, scope, or
significance beyond what the evidence warrants.

\begin{figure}[h]
\centering
\begin{tikzpicture}
\begin{axis}[
    width=0.88\linewidth, height=5cm,
    xlabel={Year},
    ylabel={Total sensational words per 100},
    xtick distance=4,
    x tick label style={rotate=45, anchor=east, font=\tiny,
                        /pgf/number format/1000 sep={}},
    ytick align=inside,
    legend columns=4,
    legend style={draw=none, fill=none, font=\scriptsize,
                  cells={anchor=west}, column sep=8pt,
                  at={(0.5,-0.28)}, anchor=north},
    tick label style={font=\tiny},
    label style={font=\small},
    enlarge x limits=0.02,
    ymin=0,
  ]
  \addplot[color={rgb,255:red,44;green,160;blue,44}, dashdotted, very thick]
    table[x=year, y=neurips, col sep=comma]
    {data/pgfplots/sensational_total.csv};
  \addlegendentry{NeurIPS}
  \addplot[color={rgb,255:red,148;green,103;blue,189}, solid, very thick]
    table[x=year, y=arxiv_ml, col sep=comma]
    {data/pgfplots/sensational_total.csv};
  \addlegendentry{arXiv ML}
  \addplot[color={rgb,255:red,140;green,86;blue,75}, dotted, thick]
    table[x=year, y=arxiv_nonml, col sep=comma]
    {data/pgfplots/sensational_total.csv};
  \addlegendentry{arXiv non-ML}
  \addplot[color={rgb,255:red,214;green,39;blue,40}, dashed, thick]
    table[x=year, y=pubmed, col sep=comma]
    {data/pgfplots/sensational_total.csv};
  \addlegendentry{PubMed}
  % ChatGPT launch marker (line only; the text label appears
  % only in fig:readability-flesch, the first chronological figure).
  \draw[black, dashed, semithick] (axis cs:2022,0) -- (axis cs:2022,30);
\end{axis}
\end{tikzpicture}
\caption{\textbf{Total sensational language per 100 abstract words,
  1987--2025.}
  The NeurIPS total rate rises by approximately 50 percent
  between 2015 and 2024. The arXiv non-ML baseline rises far
  more slowly. The divergence between ML and non-ML venues
  accelerates after the late-2022 availability of
  instruction-tuned writing assistants (vertical dashed line).
  The full 10-category breakdown appears in
  Appendix~\ref{app:figures}.}
\label{fig:sensational-total}
\end{figure}
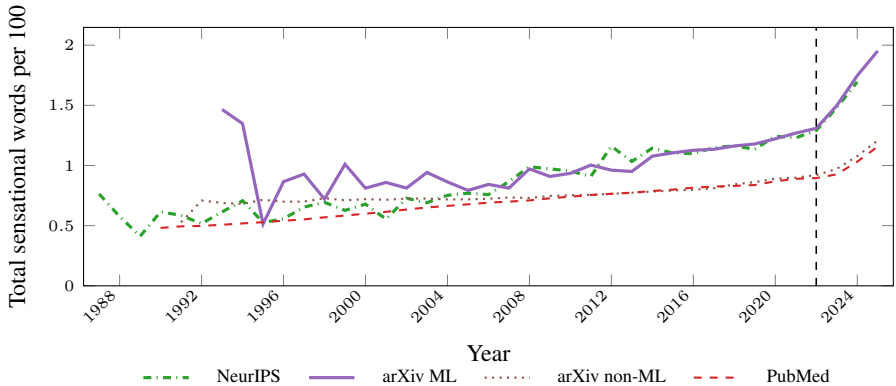

Figure~\ref{fig:sensational-total} shows total sensational
language count per 100 abstract words for NeurIPS, arXiv ML,
arXiv non-ML, and PubMed. The NeurIPS total rate rises by
approximately 50 percent between 2015 and 2024 (from 1.10 to 1.69
per 100 words). The arXiv non-ML baseline rises more
slowly over the same period. The divergence between ML and
non-ML venues accelerates after 2022, consistent with the
hypothesis that instruction-tuned writing assistants amplify
hype language rather than suppress it. The full 10-category
breakdown appears in Appendix~\ref{app:figures}.

% Total sensational language per 100 abstract words.
% Data: data/pgfplots/sensational_total.csv

\subsection{Hohmann writing style metrics}
\label{sec:hohmann}

Beyond readability formulas and hype language,
\citet{hohmann2025evolution} define a nine-metric writing style
suite that measures sentence length, syntactic complexity,
lexical density, and rhetorical stance. The suite was developed
and validated on a corpus of 20 million medical and ecological
abstracts over 70 years. The metrics and their primary literature
sources appear in Table~\ref{tab:nine} (Appendix~\ref{app:figures}).
Results for all nine metrics appear in Figure~\ref{fig:app-hohmann}
(Appendix~\ref{app:figures}).

Across the Hohmann suite, ML and non-ML venues are largely
parallel rather than divergent. Of the twelve metrics we
compute, only three (signposting, active narration, and passive
rate) show a clear widening of the ML versus non-ML gap after
2015; the remaining nine are stable or slightly converging.
This is consistent with the conclusion that the bulk of the ML
readability decline shows up in the classical readability and
sensational language families rather than in the broader Hohmann
style suite. \citet{hohmann2025evolution} note that journals
which enforce explicit writing guidelines show more stable or
improving readability profiles over the same seventy-year
period, providing empirical support for the enforcement
proposals in Section~\ref{sec:policy}.

% 4. LLM as judge.
\section{LLM as Judge Readability}
\label{sec:llm-judge}

We follow \citet{cachola2025evaluating} and apply their three
prompt templates to six open-weight instruction-tuned models on
every NeurIPS abstract from 1987 to 2024. The models are
Gemma-3-27B-Instruct, Gemma-4-31B-Instruct,
Llama-3.1-8B-Instruct, Mistral-7B-Instruct,
Mixtral-8x7B-Instruct, and Qwen2.5-32B-Instruct. Each model's
score is expressed in standard deviation units relative to its
own 1987--2022 mean, removing absolute scale differences between
models. Full prompt text and decoding parameters appear in
Appendix~\ref{app:llm-judge}.

Figure~\ref{fig:llm-judge} shows the standardized scores for all
six models under each of the three prompt variants. All six
models agree across all three prompts: LLM-judged readability is
approximately flat or gently improving over four decades, with a
notable upward shift after 2022. This result stands in direct
contrast to every classical readability metric
(Figure~\ref{fig:readability-flesch}), which shows sustained
decline over the same period.

% LLM-as-judge: model-averaged z-scores, one curve per prompt.
% Each curve is the mean of all 6 model z-scores for that prompt.
% Z-scores are standardized to each model's own 1987--2022 baseline.
% Data: data/pgfplots/llm_scores_model_avg_{prompt}.csv

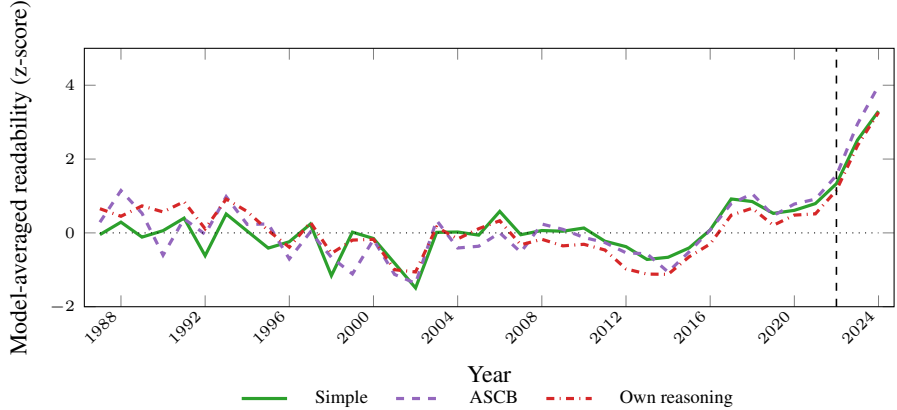
\begin{figure}[h]
\centering
\begin{tikzpicture}
\begin{axis}[
    width=0.88\linewidth, height=5cm,
    xlabel={Year},
    ylabel={Model-averaged readability (z-score)},
    xtick distance=4,
    x tick label style={rotate=45, anchor=east, font=\tiny,
                        /pgf/number format/1000 sep={}},
    ytick align=inside,
    ymin=-2, ymax=5,
    legend columns=3,
    legend style={draw=none, fill=none, font=\scriptsize,
                  cells={anchor=west}, column sep=8pt,
                  at={(0.5,-0.28)}, anchor=north},
    tick label style={font=\tiny},
    label style={font=\small},
    enlarge x limits=0.02,
  ]
  \addplot[color={rgb,255:red,44;green,160;blue,44}, solid, very thick]
    table[x=year, y=avg_z, col sep=comma]
    {data/pgfplots/llm_scores_model_avg_simple.csv};
  \addlegendentry{Simple}
  \addplot[color={rgb,255:red,148;green,103;blue,189}, dashed, very thick]
    table[x=year, y=avg_z, col sep=comma]
    {data/pgfplots/llm_scores_model_avg_ascb.csv};
  \addlegendentry{ASCB}
  \addplot[color={rgb,255:red,214;green,39;blue,40}, dashdotted, very thick]
    table[x=year, y=avg_z, col sep=comma]
    {data/pgfplots/llm_scores_model_avg_own_reasoning.csv};
  \addlegendentry{Own reasoning}
  % Baseline reference at z = 0
  \addplot[black, dotted, thin, forget plot, domain=1987:2024] {0};
  % ChatGPT launch marker (line only; the text label appears
  % only in fig:readability-flesch, the first chronological figure).
  \draw[black, dashed, semithick] (axis cs:2022,-2) -- (axis cs:2022,5);
\end{axis}
\end{tikzpicture}
\caption{\textbf{LLM as judge readability on NeurIPS abstracts,
  1987--2024.}
  Each curve is the mean z-score across all six open-weight
  models for one prompt variant, standardized to each model's
  own 1987--2022 baseline (dotted line at zero).
  All three prompts agree: LLM-judged readability is
  approximately flat before 2022 and rises sharply after the
  availability of instruction-tuned writing assistants (vertical
  dashed line), directly contradicting the classical readability
  decline in Figure~\ref{fig:readability-flesch}.}
\label{fig:llm-judge}
\end{figure}

This divergence replicates the headline finding of
\citet{cachola2025evaluating}: LLM judges underweight the
lexical and syntactic features that classical formulas penalise
and respond favourably to the surface fluency properties that
instruction-tuned writing assistants produce. The post-2022
upward shift is consistent with AI-assisted writing being
optimised for the features LLMs attend to rather than for the
working-memory costs that classical formulas measure. The
implication for policy is direct. The readability threshold in
P2 (Section~\ref{sec:policy}) cannot rely on a single LLM
judge, and must combine classical scores, Hohmann suite
measures, and calibrated human judgements.

% 5. Acronyms.
\section{Acronyms: Volume and Reuse}
\label{sec:acronyms}

\begin{figure}[h]
\centering
\pgfplotsset{
  acro style/.style={
    height=5cm, width=0.47\linewidth,
    xlabel={Year}, ylabel={Acronyms per 100 words},
    xtick distance=4, x tick label style={rotate=45, anchor=east, font=\tiny,
                                          /pgf/number format/1000 sep={}},
    ytick align=inside, ymin=0,
    tick label style={font=\tiny}, label style={font=\small},
    enlarge x limits=0.02,
  }
}
% -- title density: legend collected here ----------------------------------
\begin{tikzpicture}
\begin{axis}[acro style,
  title={\small Titles~[\textdownarrow\ better]},
  legend to name=acrolegend,
  legend columns=5,
  legend style={draw=none, fill=none, font=\scriptsize,
                cells={anchor=west}, column sep=6pt},
]
  \addplot[color={rgb,255:red,44;green,160;blue,44}, dashdotted, very thick]
    table[x=year, y=neurips, col sep=comma] {data/pgfplots/acronyms_title.csv};
  \addlegendentry{NeurIPS}
  \addplot[color={rgb,255:red,148;green,103;blue,189}, solid, very thick]
    table[x=year, y=arxiv_ml, col sep=comma] {data/pgfplots/acronyms_title.csv};
  \addlegendentry{arXiv ML}
  \addplot[color={rgb,255:red,140;green,86;blue,75}, dotted, thick]
    table[x=year, y=arxiv_nonml, col sep=comma] {data/pgfplots/acronyms_title.csv};
  \addlegendentry{arXiv non-ML}
  \addplot[color={rgb,255:red,214;green,39;blue,40}, dashed, thick]
    table[x=year, y=pubmed, col sep=comma] {data/pgfplots/acronyms_title.csv};
  \addlegendentry{PubMed}
  \addplot[black, dashed, semithick, domain=1987:2025] {2.40};
  \addlegendentry{Barnett 2020 peak (2.40)}
  \draw[black, dashed, semithick] (axis cs:2022,0) -- (axis cs:2022,100);
\end{axis}
\end{tikzpicture}%
\hfill
% -- abstract density ------------------------------------------------------
\begin{tikzpicture}
\begin{axis}[acro style, title={\small Abstracts~[\textdownarrow\ better]}]
  \addplot[color={rgb,255:red,44;green,160;blue,44}, dashdotted, very thick]
    table[x=year, y=neurips, col sep=comma] {data/pgfplots/acronyms_abstract.csv};
  \addplot[color={rgb,255:red,148;green,103;blue,189}, solid, very thick]
    table[x=year, y=arxiv_ml, col sep=comma] {data/pgfplots/acronyms_abstract.csv};
  \addplot[color={rgb,255:red,140;green,86;blue,75}, dotted, thick]
    table[x=year, y=arxiv_nonml, col sep=comma] {data/pgfplots/acronyms_abstract.csv};
  \addplot[color={rgb,255:red,214;green,39;blue,40}, dashed, thick]
    table[x=year, y=pubmed, col sep=comma] {data/pgfplots/acronyms_abstract.csv};
  \addplot[black, dashed, semithick, domain=1987:2025] {2.40};
  \draw[black, dashed, semithick] (axis cs:2022,0) -- (axis cs:2022,100);
\end{axis}
\end{tikzpicture}

\vspace{2mm}
\pgfplotslegendfromname{acrolegend}

\caption{\textbf{Acronym density in titles (left) and abstracts
  (right), 1987 to 2025.}
  Lower values indicate fewer acronyms per word (\textdownarrow\
  better). Dashed black horizontal line marks the Barnett and
  Doubleday science-wide peak of 2.40 per 100 words (2019).
  NeurIPS and arXiv ML diverge from the non-ML control around
  2017 on titles and around 2020 on abstracts.}
\label{fig:acronyms}
\end{figure}
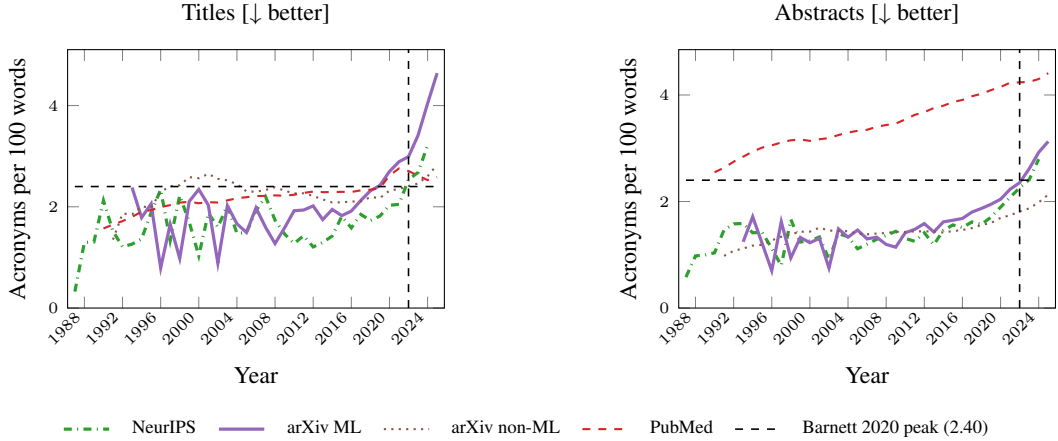

Acronyms are a direct proxy for reader memory load. Every novel
acronym occupies a working-memory slot until the reader commits
its expansion, and undefined abbreviations impair processing
fluency even when definitions are available
\citep{shulman2020jargon}.
Figure~\ref{fig:acronyms} reports acronym density (acronyms per
100 words) for titles and abstracts across NeurIPS, arXiv ML,
arXiv non-ML, and PubMed.

In titles, NeurIPS density rose from 0.33 per 100 words in 1987
to 3.21 in 2024, a tenfold increase. arXiv ML climbs from 1.60
in 2007 to 4.64 in 2025. arXiv non-ML stays between
approximately 2.1 and 2.8 across the same window, consistent
with the science-wide peak of 2.40 reported by
\citet{barnett2020growth} for 2019. PubMed titles range from
2.0 to 2.7 across the period; NeurIPS title density crossed
above PubMed only in 2024 (3.21 versus 2.53).

In abstracts, the ranking is reversed. PubMed abstract density
rises from 2.55 in 1990 to 4.30 in 2024, while NeurIPS abstract
density rises from approximately 1.0 to 2.80, so PubMed
abstracts carry roughly 1.5 times the NeurIPS rate even today.
The asymmetry between titles and abstracts is explained by the
reuse profile.
Table~\ref{tab:acronym-reuse} shows that single-use rates are
broadly comparable across venues at 29 to 32 percent, but the
2 to 9 use bucket diverges: NeurIPS and arXiv ML sit at 57 to
59 percent versus 48 to 50 percent for PubMed and the
\citet{barnett2020growth} baseline. As a consequence, 88 to 89
percent of ML acronyms appear fewer than ten times, versus 79
percent science-wide. The high-density PubMed abstract
acronyms are largely standardised vocabulary (drug names,
disease names, laboratory abbreviations, statistical reporting
such as RCT, CI, OR, HR) that appear in 10 or more papers; ML
acronyms by contrast remain mostly paper-specific. The goal of
P1 in Section~\ref{sec:policy} is to make ML acronyms behave
more like medical acronyms in reuse, not to push absolute
density below biomedical norms.

% Acronym density -- title (left) and abstract (right).
% Data: data/pgfplots/acronyms_title.csv, acronyms_abstract.csv

% acronym reuse table moved to appendix (Additional Figures)

% 6. Citations.
\section{Citations and the Fragmentation Paradox}
\label{sec:citations}

Citation count is the field's primary proxy for research impact.
If readability and citation count are positively correlated,
then the readability decline documented in
Section~\ref{sec:readability} has compounding consequences. Not
only are papers harder to read, but harder papers reach a smaller
audience and accumulate fewer citations, producing a fragmentation
dynamic within the literature. We joined citation counts from
OpenAlex and Semantic Scholar on arXiv identifier and Semantic
Scholar CorpusId. Coverage is complete for post-2010 NeurIPS.

Table~\ref{tab:citation-readability} summarises classical
readability scores and two surface linguistic measures for the
top 10 percent and bottom 10 percent cited NeurIPS papers, by
within-year citation percentile pooled over 1987 to 2024. On
twelve of fifteen reported metrics the top-cited papers score in
the easier-to-read direction. Flesch Reading Ease is 2.1 points
higher, Gunning Fog is 0.69 points lower, and LIX is 0.94 points
lower for the top decile compared to the bottom decile. Words
per sentence is 1.24 lower for top-cited papers, replicating the
direction found by \citet{hohmann2025comparing} in conservation
biology. The three exceptions, Dale-Chall, Coleman-Liau, and
FORCAST, show small reversed deltas (absolute differences below
0.3), suggesting that certain vocabulary-level formulas are less
sensitive to the stylistic differences that correlate with
citation impact in this corpus. The effects are small in absolute
terms but consistent across the majority of metrics.

% citation readability and citation LLM tables moved to appendix
% (Additional Figures)

\begin{figure}[h]
\centering
\begin{tikzpicture}
\begin{axis}[
    width=0.85\linewidth, height=5cm,
    xlabel={Year}, ylabel={Bibliography entries per paper},
    xtick distance=4,
    x tick label style={rotate=45, anchor=east, font=\tiny,
                        /pgf/number format/1000 sep={}},
    ytick align=inside,
    tick label style={font=\tiny},
    label style={font=\small},
    enlarge x limits=0.02,
  ]
  \addplot[color={rgb,255:red,44;green,160;blue,44}, dashdotted, very thick]
    table[x=year, y=references_mean, col sep=comma]
    {data/pgfplots/references_per_year.csv};
  % ChatGPT launch marker
  \draw[black, dashed, semithick] (axis cs:2022,0) -- (axis cs:2022,100);
\end{axis}
\end{tikzpicture}
\caption{\textbf{Mean bibliography length per NeurIPS paper,
  1987--2024.}
  Reference list length grew from roughly 11 entries per paper
  in 1987 to approximately 61 entries in 2022, a fivefold to
  sixfold increase over 37 years. Mean bibliography length
  declined slightly in 2023 and 2024. The cause of this reversal
  is not yet established. Vertical dashed line marks the
  late-2022 availability of instruction-tuned writing
  assistants.}
\label{fig:app-references}
\end{figure}
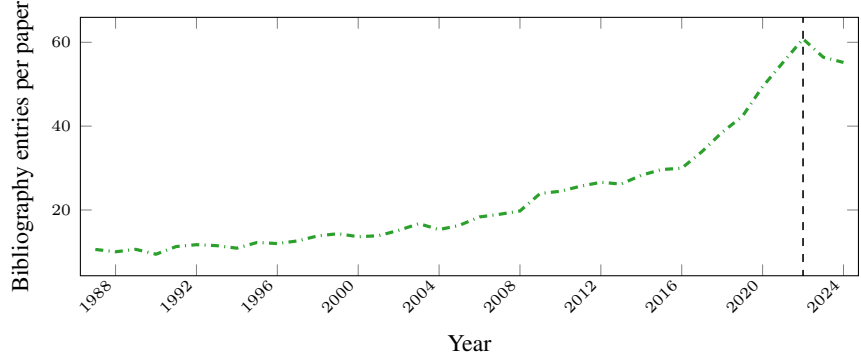

Figure~\ref{fig:app-references} shows mean bibliography length
per NeurIPS paper from 1987 to 2024. Reference list length grew
from roughly 11 entries per paper in 1987 to approximately 61
entries in 2022, a fivefold to sixfold increase over 37 years.
Mean reference list length declined slightly in 2023 and 2024
(56.4 in 2023 and 55.2 in 2024); the cause of this reversal is
not yet established and warrants further investigation. The
policy proposal in P3 (Section~\ref{sec:policy}) addresses
bibliography growth and citation quality directly. The growing
reference graph compounds the fragmentation paradox of
\citet{jeschke2019knowledge}: even as bibliographies expand, the
density of genuine readership per paper does not.

% Mean bibliography entries per NeurIPS paper, 1987-2024.
% Data: data/pgfplots/references_per_year.csv

% 7. Paper volume.
\section{The Volume Problem}
\label{sec:volume}

Writing quality does not exist in isolation. The consequences of
readability decline scale with the volume of papers that
reviewers, authors, and downstream readers must process each
year.

Figure~\ref{fig:papers} plots annual paper counts on a
logarithmic scale for NeurIPS, arXiv ML, arXiv non-ML, and
PubMed. arXiv submissions grew from roughly 300 records in 1991
to approximately 244,000 in 2024, nearly three orders of
magnitude. NeurIPS grew by a factor of
roughly 50 over the same window, from fewer than 100 accepted
papers in 1987 to approximately 4,500 in 2024 \citep{chen2025conference}.

% papers-per-year figure moved to appendix (Additional Figures)

Mean authors per accepted NeurIPS paper roughly tripled between
1987 and 2024, reflecting both genuine multi-site collaboration
and the expansion of contributorship norms
\citep{chen2025conference}. A larger author pool
implies a larger pool of competing interests over terminology,
citation practices, and writing conventions, making the
coordination problem of maintaining shared vocabulary harder
rather than easier as the field grows.

The volume problem is not merely one of reader time. It has
structural consequences for how knowledge accumulates. When the
literature grows faster than the community's capacity to
synthesise it, citation networks fragment. Papers become harder
to locate, harder to place in intellectual context, and harder
to replicate because the vocabulary that connects them is not
stable. The readability and acronym trends in
Sections~\ref{sec:readability} and~\ref{sec:acronyms} are
partially a symptom of this volume pressure.

% 8. Proposed standards.
\section{Proposed Standards for NeurIPS}
\label{sec:policy}

We propose seven standards, ordered by implementation cost. Each
standard includes a measurable success criterion evaluable one
year after a NeurIPS 2027 pilot.

\paragraph{P1. Acronym budget and approved-term list.}
No abstract may introduce more than two novel acronyms. An
acronym that appears on a venue-maintained list of approved
technical terms does not count against the budget. Every novel
acronym must be expanded on first use, and any acronym used as a
proper noun for a specific system must include a one-line
statement of the system's function. The approved list begins
with a short set of ML acronyms drawn from the community via an
open call in advance of NeurIPS 2027, and is expanded annually.
This combines the budget discipline that medical journals apply
through editorial review with the approved terminology lists
maintained by the BMJ, the NEJM, and JMLR
\citep{barnett2020growth}. Success criterion: median novel
acronym count per abstract falls by 30 percent at NeurIPS 2028
relative to the 2024 baseline.

\paragraph{P2. Readability threshold.}
A submission-time report from the writing metric calculator
(P7) flags abstracts that fall below a threshold on Flesch
Reading Ease, mean sentence length, and mean parse depth. The
threshold is soft in the first year, generating a warning
rather than a rejection, and tightens to a required revision
in subsequent years. The threshold is anchored to the NeurIPS
2022 median, preventing further degradation rather than imposing
a historically unprecedented standard. Authors who receive a
flag may include a written justification explaining why
identified passages require specialist vocabulary that cannot
be simplified without loss of precision
\citep{plaven2017readability}.
Success criterion: mean NeurIPS abstract Flesch Reading Ease
rises by five points at NeurIPS 2028.

\paragraph{P3. Stricter citation standards.}
Every paper must identify its three core citations: those
without which the paper cannot be understood and the
contribution cannot be evaluated. Every remaining citation
must support a specific claim in the text. Decorative citations
that create an impression of comprehensive scholarship without
targeting specific claims are excluded. Authors attest at
submission time that every citation in the bibliography resolves
to a retrievable paper. Success criterion: mean citations per
paper stabilises or decreases at NeurIPS 2028.

\paragraph{P4. Standalone visual elements.}
Every accepted paper should carry at least one figure or
diagram that conveys the core contribution and is
understandable without reading the full abstract. Visual
elements that are self-explanatory lower the threshold for
interdisciplinary readers and reduce the barrier to entry for
audiences outside the paper's primary subfield. The acceptance
bar for papers that communicate their claims through text
alone should be raised accordingly. Success criterion: the
proportion of accepted papers carrying a standalone explanatory
figure exceeds 80 percent at NeurIPS 2028.

\paragraph{P5. Plain language summary.}
Require a 100-word plain language summary alongside the
abstract, written for a non-specialist reader such as a
graduate student outside the paper's subfield. The summary may
not use acronyms introduced only within the paper itself. The
NEJM and the NIH public access policy both require plain
language summaries alongside technical abstracts. NeurIPS is
interdisciplinary by design and would benefit from the same
mechanism. Success criterion: 95 percent of accepted papers
carry a compliant summary, and a reader survey at NeurIPS 2028
shows higher self-reported understanding of papers outside the
respondent's primary subfield.

\paragraph{P6. Pre-registered acronym glossary.}
Every paper must submit a machine-readable glossary listing
each acronym used in the title, abstract, and main body of the
paper, with its expansion and a one-line usage note. The
glossary is a supplementary JSON or CSV file, auditable by the
writing metric calculator (P7), and visible on OpenReview as a
structured metadata field. The glossary covers the full paper
rather than only the abstract, so that reviewers and downstream
readers see the full vocabulary commitment of the work. The
glossary makes P1 enforceable at scale, supports downstream
literature search and text mining, and provides the
community-maintained approved-term list that P1 relies on.
Success criterion: 100 percent of accepted papers ship a
compliant glossary at submission time, and the community
approved-term list grows by at least 200 entries in the first
year.

\paragraph{P7. Open source audit tooling.}
Every standard proposed above is measurable, and measurement
requires tooling that is openly available, independently
verifiable, and easy to use. NeurIPS should commission or adopt
an audit tool that computes every metric in this paper from a
title and abstract, integrate it into the OpenReview submission
workflow, and publish venue-level metric dashboards each cycle.
Open source audit tooling ensures that authors can check their
own work before the review deadline, reviewers can verify
compliance, and programme chairs can monitor venue-level metric
trends over successive years. Success criterion: a venue-adopted
audit tool is integrated into the OpenReview submission workflow
at NeurIPS 2027, and venue-level metric dashboards are published
at NeurIPS 2028.

% 9. Alternative views.
% Position paper rule: optional but encouraged. Only non-strawman
% objections.
\section{Alternative Views and Counterarguments}
\label{sec:alt}

A credible position paper must engage with its strongest
objections. We address five.

\paragraph{``Acronyms are efficient for domain experts.''}
They are, within a community of practice, when the acronyms
are reused. An acronym introduced once and never cited again
is a pure reader tax: the reader pays the memory cost of
committing the expansion but receives no benefit from shared
shorthand in subsequent papers. Table~\ref{tab:acronym-reuse}
shows that 89 percent of NeurIPS acronyms are used fewer than
ten times, ten percentage points above the science-wide
baseline of \citet{barnett2020growth}. Most ML acronyms never
become shared vocabulary. P1 does not prohibit acronyms. It
asks for a budget and a reuse registry, which is precisely the
efficiency the objection seeks to preserve.

\paragraph{``This is just a consequence of field growth.''}
The arXiv non-ML categories grow at rates comparable to ML
over the same window and are drawn from the same publication
infrastructure. Those categories nevertheless sit well below ML
on every readability metric we test and at a substantially
lower acronym density (Figures~\ref{fig:readability-flesch}
and~\ref{fig:acronyms}). If field growth alone caused the
readability decline, non-ML categories should show the same
trend. They do not. Field growth is a confound we have
directly controlled for by comparing ML and non-ML arXiv
categories year by year on identical metrics.

\paragraph{``Enforced writing standards will stifle scientific creativity.''}
The BMJ, the NEJM, and JMLR have enforced explicit style
guides for decades without measurable loss of scientific
creativity or output quality. \citet{hohmann2025evolution}
find that medical journals with enforced style guidelines
show more stable or improving readability profiles compared
to journals without such enforcement. Enforcement targets the
form of communication, not its content. The seven standards in
Section~\ref{sec:policy} all operate at the level of
presentation rather than intellectual contribution.

\paragraph{``AI-assisted writing will correct these trends without intervention.''}
Current AI writing assistants optimise for surface fluency,
not for the psycholinguistic features that classical
readability formulas penalise \citep{cachola2025evaluating}.
The LLM as judge experiment in Section~\ref{sec:llm-judge}
shows directly that instruction-tuned models do not detect the
readability degradation that classical metrics reveal. They
rate the same abstracts as more readable precisely because
AI-assisted text scores highly on the surface properties they
attend to. Tooling that optimises for the correct metrics
requires deliberate design and community adoption, as
specified in P7.

\paragraph{``Stricter citation rules will penalise comprehensive scholarship.''}
The proposal in P3 is more honest than the current norm,
not less restrictive. Three declared core citations plus an
unrestricted tail of supporting citations gives reviewers a
clear signal about what is essential and what is supplementary.
A core citation is one without which the paper cannot be
understood or evaluated. Supporting citations can be any
additional references the authors find relevant. The
submission-time attestation requirement catches unresolvable
entries without requiring reviewers to verify every reference
manually.

% 10. Discussion.
\section{Discussion}
\label{sec:discussion}

The empirical picture assembled here is concrete and measurable.
NeurIPS paper volume has grown by a factor of roughly 50 since
1987, from fewer than 100 accepted papers to approximately 4,500. NeurIPS
abstracts are harder to read on every classical readability
metric we test, with the decline accelerating after 2015
relative to the arXiv non-ML control. Title acronym density on
arXiv ML exceeds the science-wide peak of
\citet{barnett2020growth} by more than 90 percent, and
88 to 89 percent of NeurIPS and arXiv ML acronyms are used fewer
than ten times, ten percentage points above the science-wide
baseline. More readable NeurIPS papers tend to receive more
citations within their venue-year cohort. LLM judges do not
detect these trends, consistent with the protocol-level findings
of \citet{cachola2025evaluating}. Taken together, these
observations constitute the preconditions for the fragmentation
paradox of \citet{jeschke2019knowledge}: information grows, but
actionable understanding does not.

Several limitations bound these findings. The analysis is
restricted to titles and abstracts. Full-text readability
signals may differ in magnitude even if not in direction
\citep{plaven2017readability,lei2016readability}. The
corpora are concentrated in English-language and
Western-institution venues. Citation analyses are restricted to
NeurIPS papers, where Semantic Scholar coverage is effectively
complete from 2010 onward. A similar audit on arXiv would require
linking arXiv identifiers to OpenAlex records at scale and is
left for future work. The LLM as judge experiment uses six
open-weight models with active parameter counts of roughly 7 to
32 billion (Mixtral-8x7B is sparse with approximately 13 billion
active parameters per token); larger closed-weight models may
recover more of the classical readability signal. The proposed standards have not yet been
evaluated prospectively, so their effects on venue-level writing
quality remain to be measured.

Four directions follow from this work. First, full-text
analysis of NeurIPS papers once licensing permits. Second,
multilingual audits of ML venues outside English. Third, a
prospective evaluation of the P1 through P7 standards at
NeurIPS 2027. Fourth, integration of the writing metric
calculator as an optional author-side tool on OpenReview before
mandatory enforcement. ML research has outgrown its
communication norms. NeurIPS has the authority and the
obligation to act, and the seven standards in
Section~\ref{sec:policy} are our concrete proposal.

\newpage

% References.
{\small
\bibliographystyle{plainnat}
\bibliography{refs}

\begin{thebibliography}{21}
\providecommand{\natexlab}[1]{#1}
\providecommand{\url}[1]{\texttt{#1}}
\expandafter\ifx\csname urlstyle\endcsname\relax
  \providecommand{\doi}[1]{doi: #1}\else
  \providecommand{\doi}{doi: \begingroup \urlstyle{rm}\Url}\fi

\bibitem[Barnett and Doubleday(2020)]{barnett2020growth}
Adrian Barnett and Zoe Doubleday.
\newblock The growth of acronyms in the scientific literature.
\newblock \emph{eLife}, 9:\penalty0 e60080, 2020.
\newblock \doi{10.7554/eLife.60080}.
\newblock URL \url{https://doi.org/10.7554/eLife.60080}.
\newblock Science wide acronym baseline: 2.4 acronyms per 100 title words
  (2019); roughly 30 percent of acronyms used once; 1.1M unique acronyms across
  24M titles and 18M abstracts (1950 to 2019).

\bibitem[Berlin(2013)]{berlin2013acronym}
Leonard Berlin.
\newblock {TAC}: {AOITROMJA}? (the acronym conundrum: advancing or impeding the
  readability of medical journal articles?).
\newblock \emph{Radiology}, 266\penalty0 (2):\penalty0 383--387, 2013.
\newblock \doi{10.1148/radiol.12121776}.
\newblock URL \url{https://doi.org/10.1148/radiol.12121776}.

\bibitem[Britton et~al.(1982)Britton, Glynn, Meyer, and
  Penland]{britton1982sentence}
Bruce~K. Britton, Shawn~M. Glynn, Bonnie J.~F. Meyer, and M.~J. Penland.
\newblock Effects of text structure on use of cognitive capacity during
  reading.
\newblock \emph{Journal of Educational Psychology}, 74\penalty0 (1):\penalty0
  51--61, 1982.
\newblock \doi{10.1037/0022-0663.74.1.51}.
\newblock URL \url{https://doi.org/10.1037/0022-0663.74.1.51}.

\bibitem[Cachola et~al.(2025)Cachola, Khashabi, and
  Dredze]{cachola2025evaluating}
Isabel Cachola, Daniel Khashabi, and Mark Dredze.
\newblock Evaluating the evaluators: Are readability metrics good measures of
  readability?
\newblock \emph{arXiv preprint arXiv:2508.19221}, 2025.
\newblock URL \url{https://arxiv.org/abs/2508.19221}.

\bibitem[Chen et~al.(2025)Chen, Duan, Lin, Wang, Wu, and
  He]{chen2025conference}
Nuo Chen, Moming Duan, Andre~Huikai Lin, Qian Wang, Jiaying Wu, and Bingsheng
  He.
\newblock Position: {T}he current {AI} conference model is unsustainable!
  diagnosing the crisis of centralized {AI} conference, 2025.
\newblock URL \url{https://arxiv.org/abs/2508.04586}.

\bibitem[Hillier et~al.(2016)Hillier, Kelly, and Klinger]{hillier2016narrative}
Ann Hillier, Ryan~P. Kelly, and Terrie Klinger.
\newblock Narrative style influences citation frequency in climate change
  science.
\newblock \emph{PLOS ONE}, 11\penalty0 (12):\penalty0 e0167983, 2016.
\newblock \doi{10.1371/journal.pone.0167983}.
\newblock URL \url{https://doi.org/10.1371/journal.pone.0167983}.

\bibitem[Hohmann and Connell(2025)]{hohmann2025comparing}
Mollie~Hawkes Hohmann and Sean~D. Connell.
\newblock Comparing the writing styles of highly and rarely cited papers in
  conservation biology.
\newblock \emph{Biological Conservation}, 307:\penalty0 111125, 2025.
\newblock \doi{10.1016/j.biocon.2025.111125}.
\newblock URL \url{https://doi.org/10.1016/j.biocon.2025.111125}.

\bibitem[Hohmann et~al.(2025)Hohmann, Barnett, King, and
  Connell]{hohmann2025evolution}
Mollie~Hawkes Hohmann, Adrian~G. Barnett, Neil King, and Sean~D. Connell.
\newblock The evolution of scientific writing: an analysis of 20 million
  abstracts over 70 years in health and medical science.
\newblock \emph{Scientometrics}, 130:\penalty0 3349--3366, 2025.
\newblock \doi{10.1007/s11192-025-05353-8}.
\newblock URL \url{https://doi.org/10.1007/s11192-025-05353-8}.
\newblock Source of the Hohmann writing style metric suite. Open source R
  implementation: \url{https://github.com/agbarnett/narrator}.

\bibitem[Jeschke et~al.(2019)Jeschke, Lokatis, Bartram, and
  Tockner]{jeschke2019knowledge}
Jonathan~M. Jeschke, Sophie Lokatis, Isabelle Bartram, and Klement Tockner.
\newblock Knowledge in the dark: scientific challenges and ways forward.
\newblock \emph{FACETS}, 4\penalty0 (1):\penalty0 423--441, 2019.
\newblock \doi{10.1139/facets-2019-0007}.
\newblock URL \url{https://doi.org/10.1139/facets-2019-0007}.

\bibitem[Lei and Yan(2016)]{lei2016readability}
Lei Lei and Su~Yan.
\newblock Readability and citations in information science: evidence from
  abstracts and articles of four journals (2003--2012).
\newblock \emph{Scientometrics}, 108\penalty0 (3):\penalty0 1155--1169, 2016.
\newblock \doi{10.1007/s11192-016-2036-9}.
\newblock URL \url{https://doi.org/10.1007/s11192-016-2036-9}.

\bibitem[Lindsay(2011)]{lindsay2011scientific}
David Lindsay.
\newblock \emph{Scientific Writing = Thinking in Words}.
\newblock CSIRO Publishing, 2011.
\newblock \doi{10.1071/9780643101579}.
\newblock URL \url{https://doi.org/10.1071/9780643101579}.

\bibitem[Mart{\'i}nez and Mammola(2021)]{martinez2021specialized}
Alejandro Mart{\'i}nez and Stefano Mammola.
\newblock Specialized terminology reduces the number of citations of scientific
  papers.
\newblock \emph{Proceedings of the Royal Society B}, 288\penalty0
  (1947):\penalty0 20202581, 2021.
\newblock \doi{10.1098/rspb.2020.2581}.
\newblock URL \url{https://doi.org/10.1098/rspb.2020.2581}.

\bibitem[Millar et~al.(2022)Millar, Batalo, and Budgell]{millar2022sensational}
Neil Millar, Bojan Batalo, and Brian Budgell.
\newblock Trends in the use of promotional language (hype) in abstracts of
  successful national institutes of health grant applications, 1985--2020.
\newblock \emph{JAMA Network Open}, 5\penalty0 (8):\penalty0 e2228676, 2022.
\newblock \doi{10.1001/jamanetworkopen.2022.28676}.
\newblock URL \url{https://doi.org/10.1001/jamanetworkopen.2022.28676}.

\bibitem[Montgomery(2003)]{montgomery2003chicago}
Scott~L. Montgomery.
\newblock \emph{The Chicago Guide to Communicating Science}.
\newblock University of Chicago Press, 2003.

\bibitem[Pinker(2014)]{pinker2014sense}
Steven Pinker.
\newblock \emph{The Sense of Style: The Thinking Person's Guide to Writing in
  the 21st Century}.
\newblock Viking, 2014.

\bibitem[Plav{\'e}n-Sigray et~al.(2017)Plav{\'e}n-Sigray, Matheson, Schiffler,
  and Thompson]{plaven2017readability}
Pontus Plav{\'e}n-Sigray, Granville~James Matheson, Bj{\"o}rn~Christian
  Schiffler, and William~Hedley Thompson.
\newblock The readability of scientific texts is decreasing over time.
\newblock \emph{eLife}, 6:\penalty0 e27725, 2017.
\newblock \doi{10.7554/eLife.27725}.
\newblock URL \url{https://doi.org/10.7554/eLife.27725}.

\bibitem[Ryba et~al.(2019)Ryba, Doubleday, and Connell]{ryba2019verbs}
Ren Ryba, Zo{\"e}~A. Doubleday, and Sean~D. Connell.
\newblock How can we boost the impact of publications? {T}ry better writing.
\newblock \emph{Proceedings of the National Academy of Sciences}, 116\penalty0
  (2):\penalty0 341--343, 2019.
\newblock \doi{10.1073/pnas.1819937116}.
\newblock URL \url{https://doi.org/10.1073/pnas.1819937116}.

\bibitem[Shulman et~al.(2020)Shulman, Dixon, Bullock, and
  Col{\'o}n~Amill]{shulman2020jargon}
Hillary~C. Shulman, Graham~N. Dixon, Odalys~M. Bullock, and Dolores
  Col{\'o}n~Amill.
\newblock The effects of jargon on processing fluency, self-perceptions, and
  scientific engagement.
\newblock \emph{Journal of Language and Social Psychology}, 39\penalty0
  (5--6):\penalty0 579--597, 2020.
\newblock \doi{10.1177/0261927X20902177}.
\newblock URL \url{https://doi.org/10.1177/0261927X20902177}.

\bibitem[Stremersch et~al.(2007)Stremersch, Verniers, and
  Verhoef]{stremersch2007numbers}
Stefan Stremersch, Isabel Verniers, and Peter~C. Verhoef.
\newblock The quest for citations: Drivers of article impact.
\newblock \emph{Journal of Marketing}, 71\penalty0 (3):\penalty0 171--193,
  2007.
\newblock \doi{10.1509/jmkg.71.3.171}.
\newblock URL \url{https://doi.org/10.1509/jmkg.71.3.171}.

\bibitem[Sweller(1994)]{sweller1994cognitive}
John Sweller.
\newblock Cognitive load theory, learning difficulty, and instructional design.
\newblock \emph{Learning and Instruction}, 4\penalty0 (4):\penalty0 295--312,
  1994.
\newblock \doi{10.1016/0959-4752(94)90003-5}.
\newblock URL \url{https://doi.org/10.1016/0959-4752(94)90003-5}.

\bibitem[Sword(2012)]{sword2012stylish}
Helen Sword.
\newblock \emph{Stylish Academic Writing}.
\newblock Harvard University Press, 2012.

\end{thebibliography}
}

\newpage

% Appendix.
\appendix

\section*{Appendix}

% A. Workflow.
\section{Workflow}
\label{app:workflow}

The analysis pipeline proceeds in four stages, summarised by
Figure~\ref{fig:workflow} in the main text.

In the first stage, raw data from each source is downloaded and
converted to a canonical Pydantic-validated schema containing
\texttt{paper\_id}, \texttt{venue}, \texttt{year},
\texttt{title}, \texttt{abstract}, \texttt{authors},
\texttt{arxiv\_primary\_category}, and
\texttt{arxiv\_categories}.

In the second stage, all per-paper metrics are computed from the
canonical schema. Classical readability metrics are computed
using the \texttt{textstat} Python library. Hohmann writing
style metrics are computed using the \texttt{en\_core\_web\_sm}
spaCy model. Acronyms are extracted using a multi-step
preprocessing pipeline following \citet{barnett2020growth}; a
token is classified as an acronym if its uppercase letter count
is at least the sum of its lowercase letter and digit counts,
and the uppercase count is at least two. LLM as judge scores
are computed for each (paper, model, prompt) triple, with three
sampled runs per triple and the median reported. Citation
counts for NeurIPS papers are retrieved from the Semantic
Scholar batch API and joined on Semantic Scholar CorpusId. arXiv
citation counts use the \texttt{ARXIV:} identifier. All
per-paper metrics are computed on titles and abstracts
independently, because titles are dense and acronym-driven while
abstracts carry the bulk of the readability signal
\citep{barnett2020growth,hohmann2025evolution}. Citation
analyses in this paper are restricted to NeurIPS, where
post-2010 coverage is effectively complete.

In the third stage, per-paper metrics are aggregated to
per-venue-year means as unweighted averages over the papers in
each bin. The arXiv ML and arXiv non-ML aggregates additionally
average across primary categories weighted by per-category paper
count, providing an ML versus non-ML contrast without a venue
confound. Year-indexed analyses cover 1991 to 2025 for arXiv,
1987 to 2024 for NeurIPS, and 1990 to 2025 for PubMed.

In the fourth stage, figures and tables are generated from the
aggregated data.

Section~\ref{sec:readability} presents the classical readability
and sensational language results.
Section~\ref{sec:llm-judge} presents the LLM as judge analysis
and its disagreement with classical metrics.
Section~\ref{sec:acronyms} covers acronym density and reuse.
Section~\ref{sec:citations} covers citation readability
correlations and bibliography growth.
Section~\ref{sec:volume} covers paper volume.
The policy proposal appears in Section~\ref{sec:policy}.

% B. Dataset.
\section{Dataset}
\label{app:data}

\paragraph{Sources and row counts.}
Table~\ref{tab:corpus} reports per-source corpus sizes at the
2026-04 snapshot. The arXiv 2025 partition is partial (snapshot
predates year end); PubMed 2025 reflects all records dated 2025
in the NLM baseline at snapshot time, including ahead-of-print
articles.

\begin{table}[h]
  \caption{Per source paper counts and coverage at the 2026-04
  snapshot. The arXiv 2025 row covers approximately three
  quarters of the calendar year.}
  \label{tab:corpus}
  \centering
  \begin{tabular}{llr}
    \toprule
    Source                 & Coverage    & Papers \\
    \midrule
    arXiv (all categories) & 1991--2025  & 2{,}823{,}623 \\
    NeurIPS                & 1987--2024  & 24{,}772 \\
    PubMed                 & 1990--2025  & 24{,}463{,}208 \\
    \bottomrule
  \end{tabular}
\end{table}

\paragraph{Acquisition.}
arXiv is processed from the
\texttt{Cornell-University/arxiv} Kaggle snapshot
(\texttt{arxiv-metadata-oai-snapshot.json}) and split into per-year
files. ML papers are defined as those whose primary category is
in \texttt{cs.LG}, \texttt{cs.AI}, \texttt{cs.CV}, \texttt{cs.CL},
\texttt{cs.NE}, \texttt{cs.RO}, \texttt{cs.IR}, \texttt{cs.MA},
\texttt{cs.HC}, \texttt{cs.CY}, or \texttt{stat.ML}.
NeurIPS is scraped from \texttt{papers.nips.cc}; pre-2008 entries
required separate abstract scraping because the early conference
data bundles full text into a single raw field. PubMed records
are downloaded from the NLM baseline FTP
(\url{https://ftp.ncbi.nlm.nih.gov/pubmed/baseline/}) as gzipped
XML files and parsed for English-language abstracts of at least
six words.

\paragraph{Canonical schema.}
Every source is normalised to the same Pydantic model:
\texttt{paper\_id}, \texttt{venue}, \texttt{year},
\texttt{title}, \texttt{abstract}, \texttt{authors},
\texttt{arxiv\_primary\_category}, \texttt{arxiv\_categories}.
All per-paper metric scripts consume only the canonical schema.

% C. LLM as Judge.
\section{LLM as Judge: Prompts and Models}
\label{app:llm-judge}

\paragraph{Prompts.}
We use the three prompt templates from \citet{cachola2025evaluating}
(their Table~10), reproduced in Table~\ref{tab:prompts}. The
instruction text of each prompt is reproduced verbatim. The only
deviation from the original is that we omit the
\texttt{Reason:~<reasoning>} output field, retaining only the
numeric score, to reduce token consumption. All three share the
same 1--5 scale, where 1 requires expert background knowledge
and 5 is readable to the general population. The \texttt{ascb}
label is taken from Cachola et al., who derived its rubric from
American Society for Cell Biology science-communication
guidelines. The prompt text itself is domain-agnostic and was
applied to NeurIPS abstracts without modification.

\begin{table}[h]
\centering
\small
\caption{The three prompt templates we apply, reproduced
verbatim from \citet{cachola2025evaluating} (Table~10) except
that the \texttt{Reason:~<reasoning>} output line is omitted.
\texttt{\{SUMMARY\}} is replaced by the abstract text at
inference.}
\label{tab:prompts}
\begin{tabular}{@{}p{\dimexpr\linewidth-2\tabcolsep\relax}@{}}
\toprule
\textbf{Simple Prompt} \\
\midrule
On a scale of 1 to 5, what is the reading ease of the following
text? 1 indicates the text requires expert background knowledge
and 5 indicates the text is readable to the general population.
Assume the reader is an adult. \texttt{\textbackslash n
\textbackslash n} \newline
Format the output as follows: \texttt{\textbackslash n} \newline
Score: \texttt{<score>} \texttt{\textbackslash n} \newline
Text: \texttt{\{SUMMARY\}} \\
\midrule
\textbf{ASCB Guidelines Prompt} \\
\midrule
On a scale of 1 to 5, what is the reading ease of the following
text? 1 indicates the text requires expert background knowledge
and 5 indicates the text is readable to the general population.
Characteristics of a highly readable text include:
\texttt{\textbackslash n} \newline
{\textendash} Know your audience, and focus and organize your
information for that particular audience.
\texttt{\textbackslash n} \newline
{\textendash} Focus on the big picture. What larger problem is
your work a part of? What major ideas or issues does your work
address? How will your work help global understanding of some
issue? \texttt{\textbackslash n} \newline
{\textendash} Avoid jargon. If you must use a technical term,
make sure to explain it, but simplify the language.
\texttt{\textbackslash n} \newline
{\textendash} Try to use metaphors or analogies to everyday
experiences that people can relate to.
\texttt{\textbackslash n} \newline
{\textendash} Underscore the importance of public support for
exploratory research and scientific information, and the role
of this information in providing the context for effective
policy making. \texttt{\textbackslash n \textbackslash n}
\newline
Assume the reader is an adult. Do not use Flesch-Kincaid or
other readability formulas. Use your own judgment to rate the
text. \texttt{\textbackslash n \textbackslash n} \newline
Format the output as follows: \texttt{\textbackslash n} \newline
Score: \texttt{<score>} \texttt{\textbackslash n} \newline
Text: \texttt{\{SUMMARY\}} \\
\midrule
\textbf{Own Reasoning Prompt} \\
\midrule
On a scale of 1 to 5, what is the reading ease of the following
text? 1 indicates the text requires expert background knowledge
and 5 indicates the text is readable to the general population.
\texttt{\textbackslash n} \newline
Assume the reader is an adult. Do not use Flesch-Kincaid or
other readability formulas. Use your own judgment to rate the
text. \texttt{\textbackslash n \textbackslash n} \newline
Format the output as follows: \texttt{\textbackslash n} \newline
Score: \texttt{<score>} \texttt{\textbackslash n} \newline
Text: \texttt{\{SUMMARY\}} \\
\bottomrule
\end{tabular}
\end{table}

\paragraph{Models and decoding.}
We run six open-weight instruction-tuned models:
Gemma-3-27B-Instruct, Gemma-4-31B-Instruct,
Llama-3.1-8B-Instruct, Mistral-7B-Instruct (v0.3),
Mixtral-8x7B-Instruct, and Qwen2.5-32B-Instruct. Decoding uses
temperature 0.7 with sampling enabled, and a cap of 32 new tokens
per generation. We run three sampled generations per
(paper, prompt, model) triple and report the median.

\paragraph{Aggregation.}
Per-paper scores are averaged across seeds. Per-year scores
are count-weighted means of per-paper medians. Scores are
standardized to the 1987--2022 per-model mean and standard
deviation before aggregation to remove absolute scale differences
between models.

% D. Additional figures and tables.
\section{Additional Figures and Tables}
\label{app:figures}

\paragraph{Acronym frequency distribution.}
Table~\ref{tab:acronym-reuse} reports the frequency distribution
of abstract acronyms by venue, referenced from
Section~\ref{sec:acronyms} in the main text.

\begin{table}[t]
\centering
\small
\caption{Acronym frequency distribution by venue (abstract acronyms). Each cell shows the percentage of unique acronyms appearing in that frequency range. Barnett and Doubleday (2020) baseline covers all scientific fields 1950--2019. The 2 to 9 use bucket is substantially larger on ML venues than on the science-wide baseline, producing an 88 to 89 percent rate of acronyms used fewer than ten times on ML venues versus 79 percent science-wide.}
\label{tab:acronym-reuse}
\begin{tabular}{lrrrrrr}
\toprule
Venue & Unique acronyms & 1 & 2--9 & 10--99 & 100--999 & 1000+ \\
\midrule
NeurIPS & 12,206 & \textbf{31.6\%} & 57.8\% & 9.7\% & 0.9\% & 0.0\% \\
arXiv ML & 115,229 & \textbf{28.9\%} & 59.1\% & 10.3\% & 1.5\% & 0.2\% \\
arXiv non-ML & 181,570 & \textbf{32.0\%} & 50.7\% & 13.7\% & 3.1\% & 0.6\% \\
PubMed & 1,692,497 & \textbf{31.1\%} & 48.5\% & 16.4\% & 3.1\% & 0.9\% \\
Barnett 2020 & 1,112,345 & 30.0\% & 49.0\% & n/a & n/a & n/a \\
\bottomrule
\end{tabular}
\end{table}

\paragraph{Citation readability and LLM scores by cohort.}
Tables~\ref{tab:citation-readability}
and~\ref{tab:citation-llm} compare classical readability and LLM
judge scores for the top 10 percent and bottom 10 percent cited
NeurIPS papers, referenced from Section~\ref{sec:citations}.

\begin{table}[t]
\centering
\small
\caption{NeurIPS readability for top-10 percent vs bottom-10 percent cited papers (within-year citation percentile, pooled 1987--2024). Values are mean with 95 percent confidence interval (normal approximation). Bold indicates the direction favours the top-cited group.}
\label{tab:citation-readability}
\begin{tabular}{llccc}
\toprule
Metric & Better & Top 10\% & Bottom 10\% & $\Delta$ (top $-$ bottom) \\
\midrule
Flesch Reading Ease & ↑ better & 20.76 $\pm$ 0.47 & 18.66 $\pm$ 0.50 & \textbf{+2.10} \\
Flesch–Kincaid Grade & ↓ better & 15.83 $\pm$ 0.09 & 16.43 $\pm$ 0.09 & \textbf{-0.60} \\
Gunning Fog Index & ↓ better & 18.86 $\pm$ 0.11 & 19.55 $\pm$ 0.11 & \textbf{-0.69} \\
SMOG Index & ↓ better & 16.64 $\pm$ 0.07 & 17.11 $\pm$ 0.08 & \textbf{-0.47} \\
Dale–Chall Score & ↓ better & 13.08 $\pm$ 0.04 & 12.95 $\pm$ 0.04 & +0.14 \\
Spache Grade & ↓ better & 7.82 $\pm$ 0.03 & 7.98 $\pm$ 0.03 & \textbf{-0.16} \\
Coleman–Liau Index & ↓ better & 16.69 $\pm$ 0.10 & 16.40 $\pm$ 0.10 & +0.29 \\
ARI & ↓ better & 17.97 $\pm$ 0.11 & 18.28 $\pm$ 0.12 & \textbf{-0.31} \\
Linsear Write & ↓ better & 15.95 $\pm$ 0.15 & 16.42 $\pm$ 0.16 & \textbf{-0.47} \\
LIX & ↓ better & 61.68 $\pm$ 0.27 & 62.62 $\pm$ 0.28 & \textbf{-0.94} \\
RIX & ↓ better & 8.70 $\pm$ 0.08 & 9.10 $\pm$ 0.09 & \textbf{-0.40} \\
FORCAST & ↓ better & 12.76 $\pm$ 0.03 & 12.69 $\pm$ 0.03 & +0.07 \\
Powers–Sumner–Kearl Grade & ↓ better & 8.32 $\pm$ 0.03 & 8.46 $\pm$ 0.03 & \textbf{-0.14} \\
Words per sentence & ↓ better & 21.99 $\pm$ 0.18 & 23.22 $\pm$ 0.20 & \textbf{-1.24} \\
Syllables per word & ↓ better & 1.94 $\pm$ 0.01 & 1.95 $\pm$ 0.01 & \textbf{-0.01} \\
\midrule
\multicolumn{5}{l}{\small $n_\text{top}=2,493$, $n_\text{bottom}=2,436$ papers.} \\
\bottomrule
\end{tabular}
\end{table}

\begin{table}[t]
\centering
\small
\caption{NeurIPS LLM as judge readability for top-10 percent vs bottom-10 percent cited papers (within-year citation percentile, pooled 1987--2024). Values are mean with 95 percent confidence interval. Score range 1 to 5; higher values indicate easier reading. Delta is top minus bottom; bold where top-cited papers score higher.}
\label{tab:citation-llm}
\begin{tabular}{llccc}
\toprule
Prompt & Model & Top 10\% & Bottom 10\% & $\Delta$ \\
\midrule
\multicolumn{5}{l}{\textbf{Simple}} \\
 & Gemma-3-27B & 2.24 $\pm$ 0.02 & 2.08 $\pm$ 0.01 & \textbf{+0.16} \\
 & Gemma-4-31B & 1.80 $\pm$ 0.02 & 1.47 $\pm$ 0.02 & \textbf{+0.34} \\
 & Llama-3.1-8B & 2.25 $\pm$ 0.02 & 1.99 $\pm$ 0.02 & \textbf{+0.26} \\
 & Mistral-7B & 3.11 $\pm$ 0.02 & 2.72 $\pm$ 0.03 & \textbf{+0.39} \\
 & Mixtral-8x7B & 3.35 $\pm$ 0.03 & 2.91 $\pm$ 0.03 & \textbf{+0.44} \\
 & Qwen2.5-32B & 2.19 $\pm$ 0.02 & 1.96 $\pm$ 0.02 & \textbf{+0.22} \\
\midrule
\multicolumn{5}{l}{\textbf{ASCB}} \\
 & Gemma-3-27B & 2.15 $\pm$ 0.01 & 2.04 $\pm$ 0.01 & \textbf{+0.11} \\
 & Gemma-4-31B & 1.57 $\pm$ 0.02 & 1.29 $\pm$ 0.02 & \textbf{+0.29} \\
 & Llama-3.1-8B & 1.73 $\pm$ 0.02 & 1.48 $\pm$ 0.02 & \textbf{+0.25} \\
 & Mistral-7B & 3.85 $\pm$ 0.03 & 3.37 $\pm$ 0.04 & \textbf{+0.48} \\
 & Mixtral-8x7B & 3.66 $\pm$ 0.02 & 3.32 $\pm$ 0.02 & \textbf{+0.34} \\
 & Qwen2.5-32B & 1.89 $\pm$ 0.01 & 1.67 $\pm$ 0.02 & \textbf{+0.22} \\
\midrule
\multicolumn{5}{l}{\textbf{Own reasoning}} \\
 & Gemma-3-27B & 2.30 $\pm$ 0.02 & 2.11 $\pm$ 0.01 & \textbf{+0.19} \\
 & Gemma-4-31B & 1.76 $\pm$ 0.02 & 1.42 $\pm$ 0.02 & \textbf{+0.33} \\
 & Llama-3.1-8B & 2.38 $\pm$ 0.02 & 2.13 $\pm$ 0.02 & \textbf{+0.25} \\
 & Mistral-7B & 3.54 $\pm$ 0.03 & 3.17 $\pm$ 0.03 & \textbf{+0.37} \\
 & Mixtral-8x7B & 3.47 $\pm$ 0.02 & 3.02 $\pm$ 0.03 & \textbf{+0.45} \\
 & Qwen2.5-32B & 2.33 $\pm$ 0.02 & 2.06 $\pm$ 0.02 & \textbf{+0.27} \\
\midrule
\multicolumn{5}{l}{\small $n_\text{top}\approx2,385$, $n_\text{bottom}\approx2,283$ papers per model per prompt.} \\
\bottomrule
\end{tabular}
\end{table}

\begin{figure}[h]
\centering
\begin{tikzpicture}
\begin{axis}[
    width=0.88\linewidth, height=5cm,
    xlabel={Year}, ylabel={Papers per year (log scale)},
    ymode=log,
    xtick distance=4,
    x tick label style={rotate=45, anchor=east, font=\tiny,
                        /pgf/number format/1000 sep={}},
    ytick align=inside,
    legend columns=4,
    legend style={draw=none, fill=none, font=\scriptsize,
                  cells={anchor=west}, column sep=8pt,
                  at={(0.5,-0.28)}, anchor=north},
    tick label style={font=\tiny}, label style={font=\small},
    enlarge x limits=0.02,
  ]
  \addplot[color={rgb,255:red,44;green,160;blue,44}, dashdotted, very thick]
    table[x=year, y=neurips, col sep=comma] {data/pgfplots/papers_per_year.csv};
  \addlegendentry{NeurIPS}
  \addplot[color={rgb,255:red,148;green,103;blue,189}, solid, very thick]
    table[x=year, y=arxiv_ml, col sep=comma] {data/pgfplots/papers_per_year.csv};
  \addlegendentry{arXiv ML}
  \addplot[color={rgb,255:red,140;green,86;blue,75}, dotted, thick]
    table[x=year, y=arxiv_nonml, col sep=comma] {data/pgfplots/papers_per_year.csv};
  \addlegendentry{arXiv non-ML}
  \addplot[color={rgb,255:red,214;green,39;blue,40}, dashed, thick]
    table[x=year, y=pubmed, col sep=comma] {data/pgfplots/papers_per_year.csv};
  \addlegendentry{PubMed}
  \draw[black, dashed, semithick] (axis cs:2022,1) -- (axis cs:2022,1e8);
\end{axis}
\end{tikzpicture}

\vspace{2mm}

\caption{\textbf{Papers per year, 1987 to 2024.}
  arXiv ML submissions grew by roughly three orders of magnitude
  between 1992 and 2024. NeurIPS grew by a factor of roughly 50
  over the same window, from fewer than 100 accepted papers in
  1987 to over 4,500 in 2024. PubMed and arXiv non-ML show
  sustained but slower growth over the same period. The volume
  increase is the structural backdrop against which the
  readability and acronym trends in
  Sections~\ref{sec:readability} and~\ref{sec:acronyms} must
  be interpreted.}
\label{fig:papers}
\end{figure}
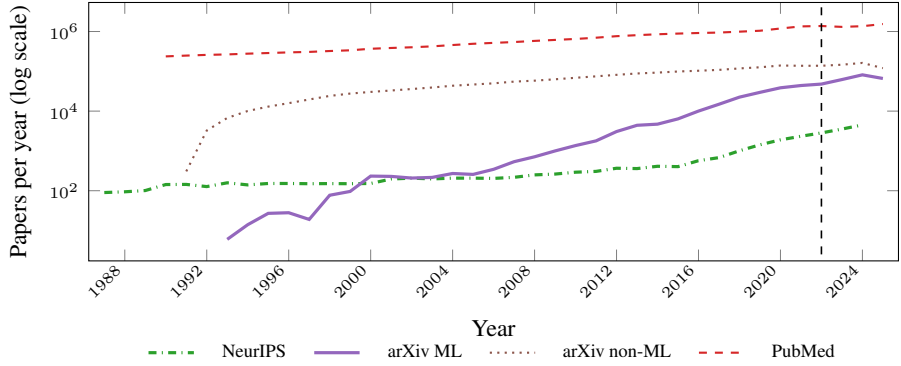

\paragraph{Annual paper volume.}
Figure~\ref{fig:papers} plots annual paper counts for NeurIPS,
arXiv ML, arXiv non-ML, and PubMed, referenced from
Section~\ref{sec:volume}.

% Papers per year on a log scale.
% Data: data/pgfplots/papers_per_year.csv

\paragraph{Writing style metrics: sources and definitions.}

\begin{table}[h]
  \caption{Writing style metrics audited per paper, with the
  original sources cited by \citet{hohmann2025evolution}.}
  \label{tab:nine}
  \centering
  \begin{tabular}{ll}
    \toprule
    Metric & Primary sources \\
    \midrule
    Sentence length      & \citet{britton1982sentence} \\
    Sign posting         & \citet{lindsay2011scientific, montgomery2003chicago} \\
    Hedging              & \citet{montgomery2003chicago, pinker2014sense, sword2012stylish} \\
    Noun chunks          & \citet{lindsay2011scientific, montgomery2003chicago, pinker2014sense} \\
    Active narration     & \citet{hillier2016narrative, pinker2014sense, sword2012stylish} \\
    Verbs                & \citet{hillier2016narrative, ryba2019verbs} \\
    Nouns                & \citet{hohmann2025evolution} \\
    Numbers              & \citet{stremersch2007numbers} \\
    Sensational language & \citet{millar2022sensational} \\
    \bottomrule
  \end{tabular}
\end{table}

\paragraph{Full classical readability metrics.}
Figures~\ref{fig:app-readability} and~\ref{fig:app-readability-b}
together show all 15 classical readability metrics computed on
abstracts of NeurIPS, arXiv ML, arXiv non-ML, and PubMed. In every
case the direction is consistent with
Figure~\ref{fig:readability-flesch}: ML venues diverge from the
non-ML baseline after approximately 2015 and accelerate after 2020.

% Classical readability — appendix, split into two 3-column figures
% to fit within the NeurIPS text width.
% Data: data/pgfplots/readability_<metric>.csv

\providecommand{\readpanel}[3]{%
  \nextgroupplot[title={\tiny #1~[#3]},
                 title style={font=\tiny, yshift=-1.5mm}]
  \addplot[color={rgb,255:red,44;green,160;blue,44}, dashdotted, very thick]
    table[x=year, y=neurips, col sep=comma]{data/pgfplots/#2.csv};
  \addplot[color={rgb,255:red,148;green,103;blue,189}, solid, thick]
    table[x=year, y=arxiv_ml, col sep=comma]{data/pgfplots/#2.csv};
  \addplot[color={rgb,255:red,140;green,86;blue,75}, dotted, thick]
    table[x=year, y=arxiv_nonml, col sep=comma]{data/pgfplots/#2.csv};
  \addplot[color={rgb,255:red,214;green,39;blue,40}, dashed, thick]
    table[x=year, y=pubmed, col sep=comma]{data/pgfplots/#2.csv};
  \draw[black, dashed, semithick]
    ({axis cs:2022,0} |- {rel axis cs:0,0})
    -- ({axis cs:2022,0} |- {rel axis cs:0,1});
}

% ── Figure A: 8 readability metrics, 3 by 3 layout (1 blank slot) ────────────
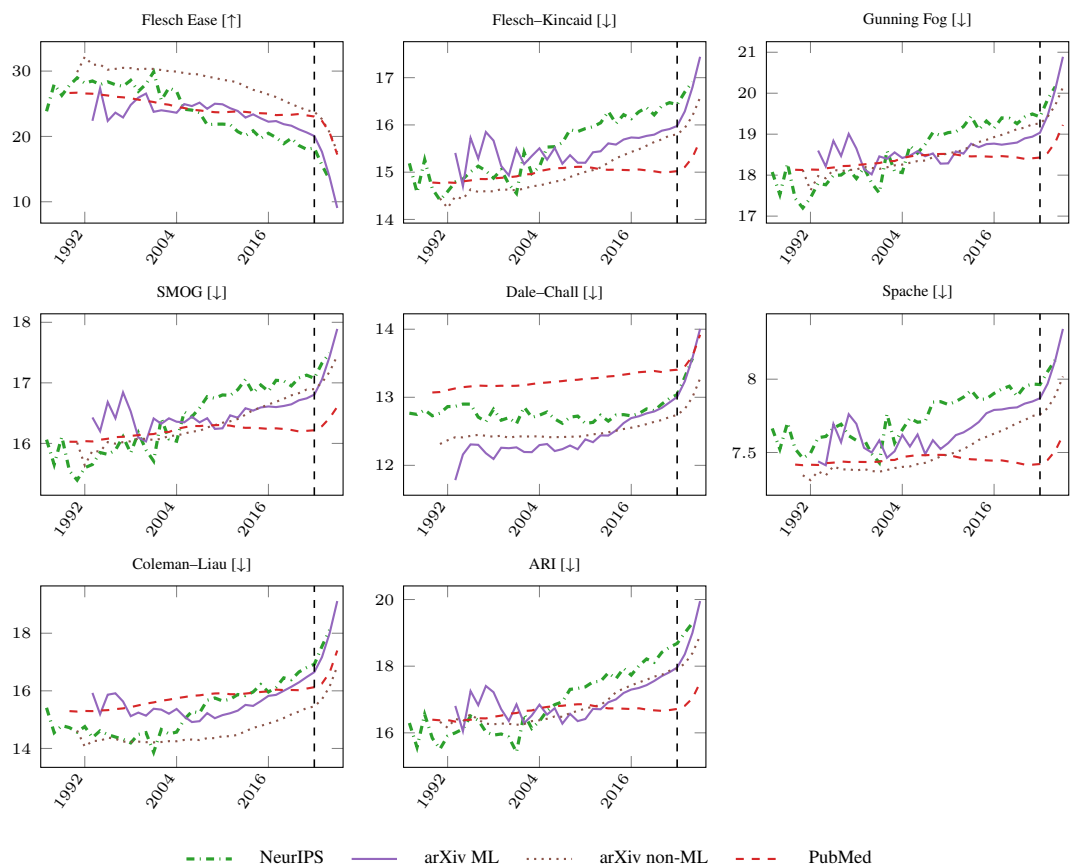
\begin{figure}[htbp]
\centering
\begin{tikzpicture}
\begin{groupplot}[
    group style={
      group size=3 by 3,
      horizontal sep=8mm,
      vertical sep=12mm,
    },
    width=40mm, height=2.4cm,
    scale only axis,
    xlabel={}, ylabel={},
    xtick distance=12,
    x tick label style={rotate=55, anchor=east, font=\tiny,
                        /pgf/number format/1000 sep={}},
    y tick label style={font=\tiny},
    ytick align=inside,
    enlarge x limits=0.02,
    tick label style={font=\tiny},
  ]
  % First panel: legend collected here
  \nextgroupplot[
    title={\tiny Flesch Ease~[\textuparrow]},
    title style={font=\tiny, yshift=-1.5mm},
    legend to name=readlegend,
    legend columns=4,
    legend style={draw=none, fill=none, font=\scriptsize,
                  cells={anchor=west}, column sep=8pt},
  ]
  \addplot[color={rgb,255:red,44;green,160;blue,44}, dashdotted, very thick]
    table[x=year, y=neurips, col sep=comma]{data/pgfplots/readability_flesch_ease.csv};
  \addlegendentry{NeurIPS}
  \addplot[color={rgb,255:red,148;green,103;blue,189}, solid, thick]
    table[x=year, y=arxiv_ml, col sep=comma]{data/pgfplots/readability_flesch_ease.csv};
  \addlegendentry{arXiv ML}
  \addplot[color={rgb,255:red,140;green,86;blue,75}, dotted, thick]
    table[x=year, y=arxiv_nonml, col sep=comma]{data/pgfplots/readability_flesch_ease.csv};
  \addlegendentry{arXiv non-ML}
  \addplot[color={rgb,255:red,214;green,39;blue,40}, dashed, thick]
    table[x=year, y=pubmed, col sep=comma]{data/pgfplots/readability_flesch_ease.csv};
  \addlegendentry{PubMed}
  \draw[black, dashed, semithick]
    ({axis cs:2022,0} |- {rel axis cs:0,0})
    -- ({axis cs:2022,0} |- {rel axis cs:0,1});
  % Remaining 7 panels of part A
  \readpanel{Flesch--Kincaid}{readability_flesch_kincaid}{\textdownarrow}
  \readpanel{Gunning Fog}{readability_gunning_fog}{\textdownarrow}
  \readpanel{SMOG}{readability_smog}{\textdownarrow}
  \readpanel{Dale--Chall}{readability_dale_chall}{\textdownarrow}
  \readpanel{Spache}{readability_spache}{\textdownarrow}
  \readpanel{Coleman--Liau}{readability_coleman_liau}{\textdownarrow}
  \readpanel{ARI}{readability_ari}{\textdownarrow}
  % 9th slot blank for layout
  \nextgroupplot[hide axis, scale only axis, width=40mm, height=2.4cm]
\end{groupplot}
\end{tikzpicture}

\vspace{2mm}
\pgfplotslegendfromname{readlegend}

\caption{\textbf{Classical readability metrics, part A: Flesch
  family and grade-level formulas, 1987--2025.}
  Arrows indicate the direction of easier reading
  (\textuparrow\ = higher is better; \textdownarrow\ = lower is
  better). Every grade-level metric rises on NeurIPS and arXiv
  ML after 2015 and accelerates after 2020; Flesch Reading Ease
  drops correspondingly. Vertical dashed lines mark the
  late-2022 availability of instruction-tuned writing
  assistants. Part B with the remaining seven metrics appears in
  Figure~\ref{fig:app-readability-b}.}
\label{fig:app-readability}
\end{figure}

% ── Figure B: 7 readability metrics, 3 by 3 layout (2 blank slots) ───────────
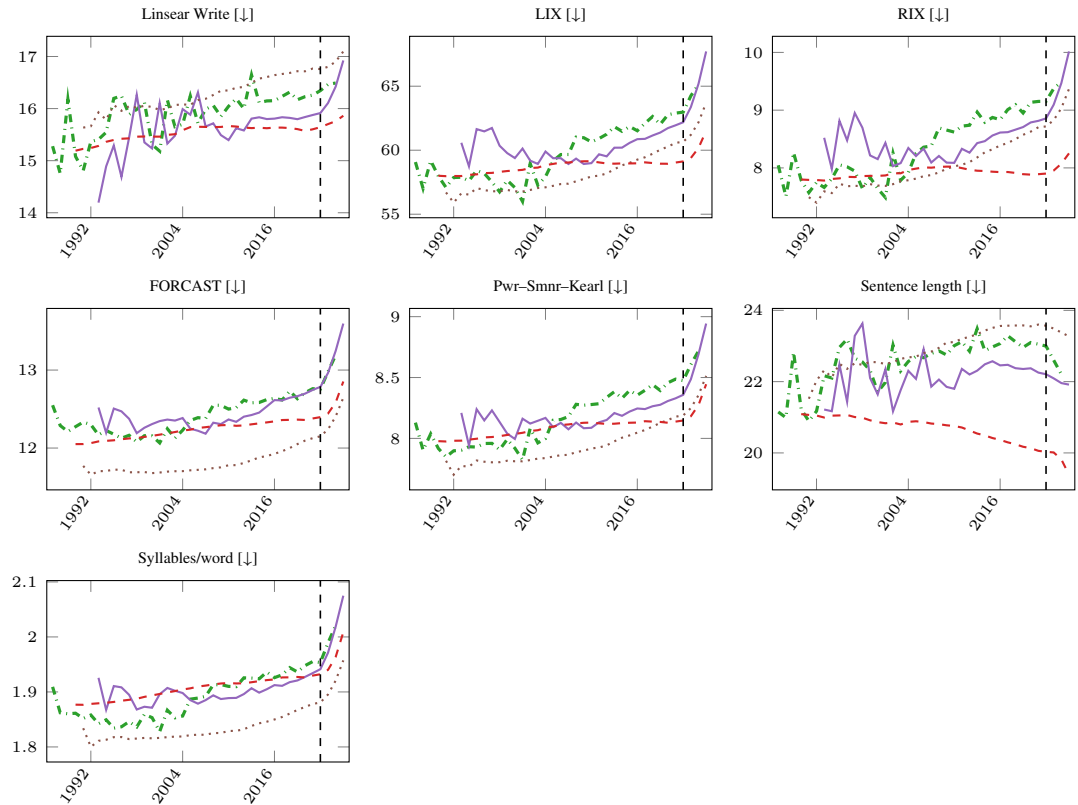
\begin{figure}[htbp]
\centering
\begin{tikzpicture}
\begin{groupplot}[
    group style={
      group size=3 by 3,
      horizontal sep=8mm,
      vertical sep=12mm,
    },
    width=40mm, height=2.4cm,
    scale only axis,
    xlabel={}, ylabel={},
    xtick distance=12,
    x tick label style={rotate=55, anchor=east, font=\tiny,
                        /pgf/number format/1000 sep={}},
    y tick label style={font=\tiny},
    ytick align=inside,
    enlarge x limits=0.02,
    tick label style={font=\tiny},
  ]
  \readpanel{Linsear Write}{readability_linsear_write}{\textdownarrow}
  \readpanel{LIX}{readability_lix}{\textdownarrow}
  \readpanel{RIX}{readability_rix}{\textdownarrow}
  \readpanel{FORCAST}{readability_forcast}{\textdownarrow}
  \readpanel{Pwr--Smnr--Kearl}{readability_powers_sumner_kearl}{\textdownarrow}
  \readpanel{Sentence length}{readability_avg_sentence_length}{\textdownarrow}
  \readpanel{Syllables/word}{readability_avg_syllables_per_word}{\textdownarrow}
  % Two blank slots for layout
  \nextgroupplot[hide axis, scale only axis, width=40mm, height=2.4cm]
  \nextgroupplot[hide axis, scale only axis, width=40mm, height=2.4cm]
\end{groupplot}
\end{tikzpicture}

\caption{\textbf{Classical readability metrics, part B: Linsear
  Write, LIX, RIX, FORCAST, Powers--Sumner--Kearl, sentence
  length, and syllables per word, 1987--2025.}
  Arrows indicate the direction of easier reading
  (\textdownarrow\ = lower is better). Vertical dashed lines mark
  the late-2022 availability of instruction-tuned writing
  assistants. Part A with the Flesch family and the first eight
  grade-level formulas appears in
  Figure~\ref{fig:app-readability}.}
\label{fig:app-readability-b}
\end{figure}

\paragraph{Hohmann writing style metrics.}
Figure~\ref{fig:app-hohmann} shows all nine Hohmann writing
style metrics and four additional linguistic complexity measures
across all venues. In every metric, ML venues diverge from the
arXiv non-ML baseline after approximately 2015.

% Hohmann writing style metrics on NeurIPS, arXiv ML, arXiv non-ML, PubMed.
% Data: data/pgfplots/<metric>.csv
% Legend collected from first panel and rendered below, matching fig_readability.

\providecommand{\hohmannpanel}[2]{%
  \nextgroupplot[title={\tiny #1},
                 title style={font=\tiny, yshift=-1.5mm}]
  \addplot[color={rgb,255:red,44;green,160;blue,44}, dashdotted, very thick]
    table[x=year, y=neurips, col sep=comma]{data/pgfplots/#2.csv};
  \addplot[color={rgb,255:red,148;green,103;blue,189}, solid, thick]
    table[x=year, y=arxiv_ml, col sep=comma]{data/pgfplots/#2.csv};
  \addplot[color={rgb,255:red,140;green,86;blue,75}, dotted, thick]
    table[x=year, y=arxiv_nonml, col sep=comma]{data/pgfplots/#2.csv};
  \addplot[color={rgb,255:red,214;green,39;blue,40}, dashed, thick]
    table[x=year, y=pubmed, col sep=comma]{data/pgfplots/#2.csv};
  \draw[black, dashed, semithick]
    ({axis cs:2022,0} |- {rel axis cs:0,0})
    -- ({axis cs:2022,0} |- {rel axis cs:0,1});
}

\begin{figure}[htbp]
\centering
\begin{tikzpicture}
\begin{groupplot}[
    group style={
      group size=3 by 4,
      horizontal sep=8mm,
      vertical sep=12mm,
    },
    width=40mm, height=2.4cm,
    scale only axis,
    xlabel={}, ylabel={},
    xtick distance=12,
    x tick label style={rotate=55, anchor=east, font=\tiny,
                        /pgf/number format/1000 sep={}},
    y tick label style={font=\tiny},
    ytick align=inside,
    enlarge x limits=0.02,
    tick label style={font=\tiny},
  ]
  % First panel: legend collected here
  \nextgroupplot[
    title={\tiny Sentence length},
    title style={font=\tiny, yshift=-1.5mm},
    legend to name=hohmannlegend,
    legend columns=4,
    legend style={draw=none, fill=none, font=\scriptsize,
                  cells={anchor=west}, column sep=8pt},
  ]
  \addplot[color={rgb,255:red,44;green,160;blue,44}, dashdotted, very thick]
    table[x=year, y=neurips, col sep=comma]{data/pgfplots/sentence_length.csv};
  \addlegendentry{NeurIPS}
  \addplot[color={rgb,255:red,148;green,103;blue,189}, solid, thick]
    table[x=year, y=arxiv_ml, col sep=comma]{data/pgfplots/sentence_length.csv};
  \addlegendentry{arXiv ML}
  \addplot[color={rgb,255:red,140;green,86;blue,75}, dotted, thick]
    table[x=year, y=arxiv_nonml, col sep=comma]{data/pgfplots/sentence_length.csv};
  \addlegendentry{arXiv non-ML}
  \addplot[color={rgb,255:red,214;green,39;blue,40}, dashed, thick]
    table[x=year, y=pubmed, col sep=comma]{data/pgfplots/sentence_length.csv};
  \addlegendentry{PubMed}
  \draw[black, dashed, semithick]
    ({axis cs:2022,0} |- {rel axis cs:0,0})
    -- ({axis cs:2022,0} |- {rel axis cs:0,1});
  % Remaining 11 panels via macro
  \hohmannpanel{Parse depth}{parse_depth}
  \hohmannpanel{NP density}{np_density}
  \hohmannpanel{Noun chunks/100}{noun_chunks}
  \hohmannpanel{Nouns/100}{nouns}
  \hohmannpanel{Verbs/100}{verbs}
  \hohmannpanel{Numbers/100}{numbers}
  \hohmannpanel{Signposting/100}{signposting}
  \hohmannpanel{Hedging/100}{hedging}
  \hohmannpanel{Active narration}{active_narration}
  \hohmannpanel{Passive rate}{passive_rate}
  \hohmannpanel{Type-token ratio}{ttr}
\end{groupplot}
\end{tikzpicture}

\vspace{2mm}
\pgfplotslegendfromname{hohmannlegend}

\caption{\textbf{Hohmann writing style metrics on NeurIPS, arXiv ML,
  arXiv non-ML, and PubMed, 1987--2025.}
  PubMed data are available only for the four metrics computed
  by lightweight regex and word-level rules (sentence length,
  signposting, hedging, and active narration). The remaining
  metrics depend on full spaCy dependency parsing and were not
  run on the 24.5 million paper PubMed corpus because of compute
  cost; the corresponding panels show NeurIPS, arXiv ML, and
  arXiv non-ML only. Across the suite, ML and non-ML venues are
  largely parallel; only signposting, active narration, and
  passive rate show a clear widening of the ML versus non-ML
  gap after 2015.}
\label{fig:app-hohmann}
\end{figure}
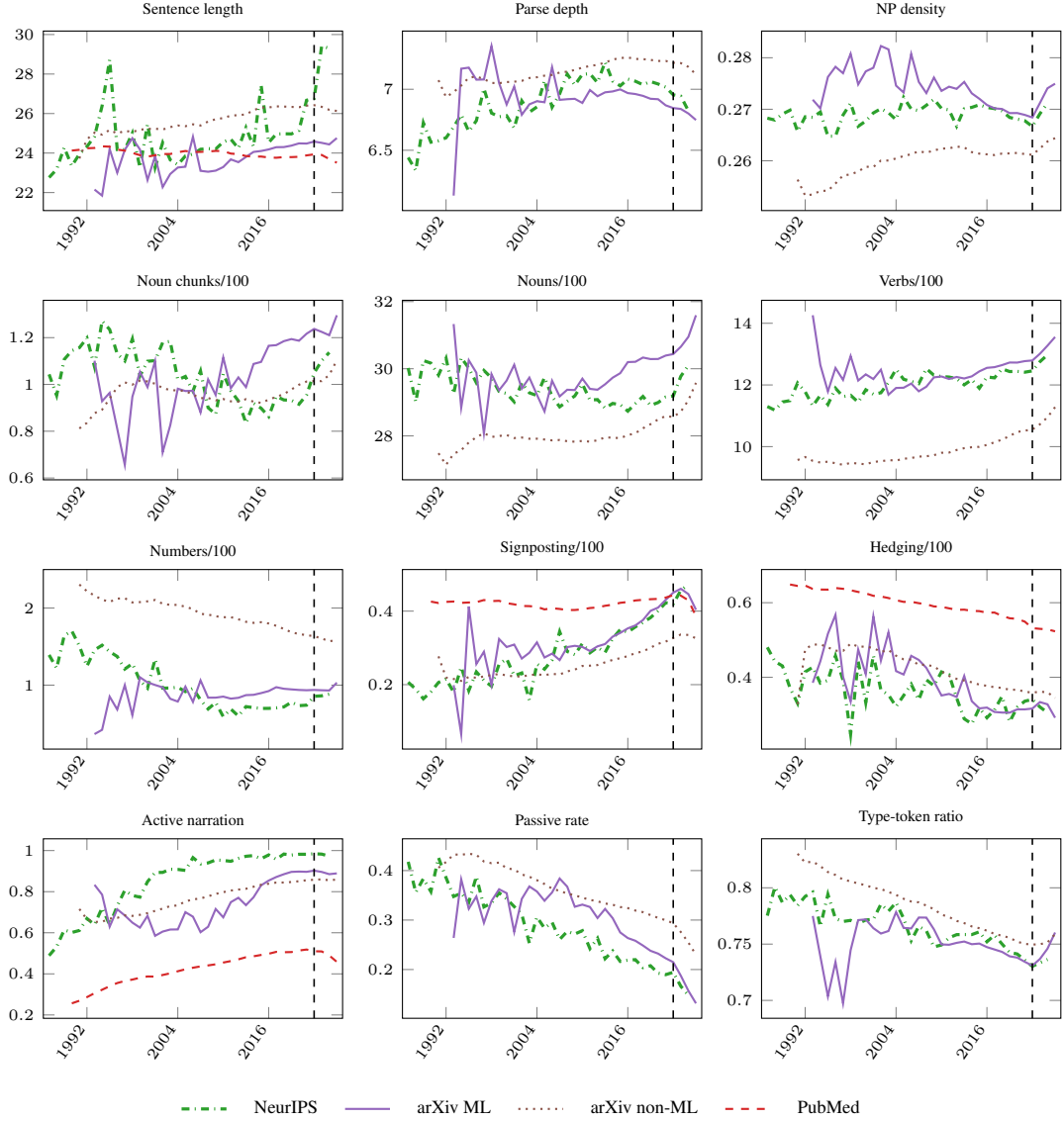

\paragraph{Full sensational language breakdown.}
Figure~\ref{fig:app-sensational} shows all ten categories of
sensational language per 100 abstract words. Novelty and scale
rise sharply on ML venues after 2015 (roughly 2.2 times and 3.0
times their 2015 levels by 2024); the total rate and rigor rise
more modestly over the same period.

% Full 10-panel sensational language figure (appendix).
% Data: data/pgfplots/sensational_<category>.csv

\providecommand{\senspanel}[2]{%
  \nextgroupplot[title={\tiny #1},
                 title style={font=\tiny, yshift=-1.5mm}]
  \addplot[color={rgb,255:red,44;green,160;blue,44}, dashdotted, very thick]
    table[x=year, y=neurips, col sep=comma]{data/pgfplots/#2.csv};
  \addplot[color={rgb,255:red,148;green,103;blue,189}, solid, thick]
    table[x=year, y=arxiv_ml, col sep=comma]{data/pgfplots/#2.csv};
  \addplot[color={rgb,255:red,140;green,86;blue,75}, dotted, thick]
    table[x=year, y=arxiv_nonml, col sep=comma]{data/pgfplots/#2.csv};
  \addplot[color={rgb,255:red,214;green,39;blue,40}, dashed, thick]
    table[x=year, y=pubmed, col sep=comma]{data/pgfplots/#2.csv};
  \draw[black, dashed, semithick]
    ({axis cs:2022,0} |- {rel axis cs:0,0})
    -- ({axis cs:2022,0} |- {rel axis cs:0,1});
}

\begin{figure}[htbp]
\centering
\begin{tikzpicture}
\begin{groupplot}[
    group style={
      group size=3 by 4,
      horizontal sep=8mm,
      vertical sep=12mm,
    },
    width=40mm, height=2.4cm,
    scale only axis,
    xlabel={}, ylabel={},
    xtick distance=12,
    x tick label style={rotate=55, anchor=east, font=\tiny,
                        /pgf/number format/1000 sep={}},
    y tick label style={font=\tiny},
    ytick align=inside,
    enlarge x limits=0.02,
    ymin=0,
  ]
  % First panel: legend collected here
  \nextgroupplot[
    title={\tiny Importance},
    title style={font=\tiny, yshift=-1.5mm},
    legend to name=senslegend,
    legend columns=4,
    legend style={draw=none, fill=none, font=\scriptsize,
                  cells={anchor=west}, column sep=8pt},
  ]
  \addplot[color={rgb,255:red,44;green,160;blue,44}, dashdotted, very thick]
    table[x=year, y=neurips, col sep=comma]{data/pgfplots/sensational_importance.csv};
  \addlegendentry{NeurIPS}
  \addplot[color={rgb,255:red,148;green,103;blue,189}, solid, thick]
    table[x=year, y=arxiv_ml, col sep=comma]{data/pgfplots/sensational_importance.csv};
  \addlegendentry{arXiv ML}
  \addplot[color={rgb,255:red,140;green,86;blue,75}, dotted, thick]
    table[x=year, y=arxiv_nonml, col sep=comma]{data/pgfplots/sensational_importance.csv};
  \addlegendentry{arXiv non-ML}
  \addplot[color={rgb,255:red,214;green,39;blue,40}, dashed, thick]
    table[x=year, y=pubmed, col sep=comma]{data/pgfplots/sensational_importance.csv};
  \addlegendentry{PubMed}
  \draw[black, dashed, semithick]
    ({axis cs:2022,0} |- {rel axis cs:0,0})
    -- ({axis cs:2022,0} |- {rel axis cs:0,1});
  % Remaining 9 panels via macro
  \senspanel{Novelty}{sensational_novelty}
  \senspanel{Rigor}{sensational_rigor}
  \senspanel{Scale}{sensational_scale}
  \senspanel{Utility}{sensational_utility}
  \senspanel{Quality}{sensational_quality}
  \senspanel{Attitude}{sensational_attitude}
  \senspanel{Problem}{sensational_problem}
  \senspanel{Additional}{sensational_additional}
  \senspanel{Total}{sensational_total}
  \nextgroupplot[hide axis, scale only axis, width=40mm, height=2.4cm]
  \nextgroupplot[hide axis, scale only axis, width=40mm, height=2.4cm]
\end{groupplot}
\end{tikzpicture}

\vspace{2mm}
\pgfplotslegendfromname{senslegend}

\caption{\textbf{All 10 sensational language categories per 100
  abstract words, 1987--2025.}
  Novelty and scale rise sharply on NeurIPS and arXiv ML after
  2015 (roughly 2.2 times and 3.0 times their 2015 levels by
  2024); the total rate and rigor rise more modestly. Vertical
  dashed lines mark the late-2022 availability of
  instruction-tuned writing assistants.}
\label{fig:app-sensational}
\end{figure}
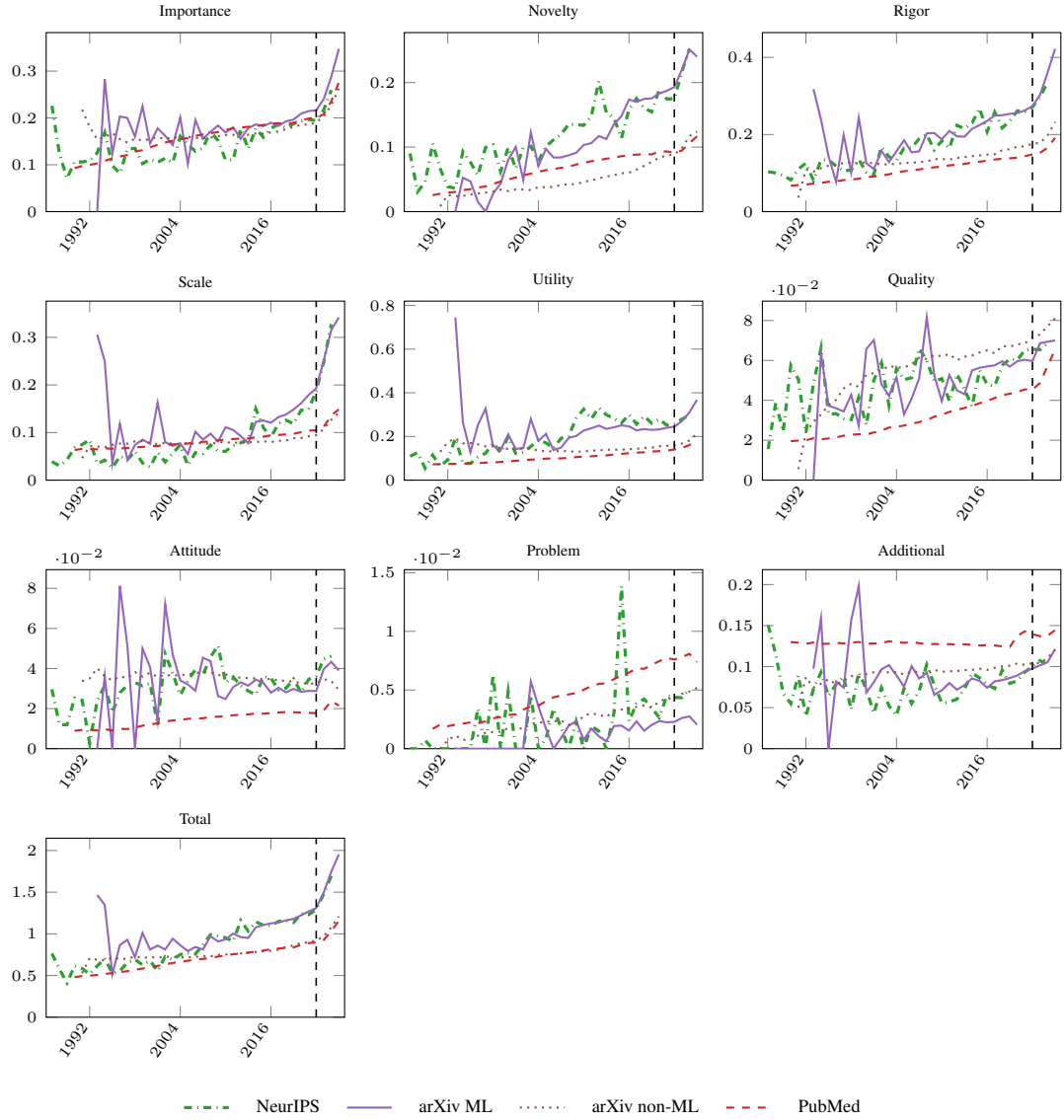

% E. Metric audit.
\section{Metric Audit of This Paper}
\label{app:self-audit}
To substantiate the claim that ML writing can meet the proposed
standards, we apply the same scripts that produced
Sections~\ref{sec:readability} through~\ref{sec:acronyms} to
this paper's own title and abstract
(Table~\ref{tab:self-audit}). On acronym budget we are within
P1: the abstract uses two distinct acronyms, NeurIPS and LLM,
both of which are standardised community terms suitable for the
approved-term list and so do not count as novel coinages. On
readability the abstract sits well above the NeurIPS abstract
distribution on Flesch Reading Ease (93rd percentile) and below
the median on Flesch-Kincaid grade level (13th percentile),
which is the direction P2 is designed to encourage.

\begin{table}[h]
  \caption{Metric self-audit of this manuscript. Numbers are
  produced by running the same per-paper metric scripts used
  throughout the paper on this paper's own title and abstract.
  Percentile refers to the NeurIPS 1987--2024 abstract
  distribution for the same metric (n=24,772). Hohmann-suite and
  sensational-language metrics are computed at the abstract level
  only.}
  \label{tab:self-audit}
  \centering
  \small
  \begin{tabular}{lrrr}
    \toprule
    Metric & Title & Abstract & Percentile \\
    \midrule
    \multicolumn{4}{l}{\textit{Classical readability (15 metrics)}} \\
    \midrule
    Flesch Reading Ease       & 53.66 & 36.62 & 93.0\% \\
    Flesch--Kincaid Grade     & 8.76  & 13.80 & 13.1\% \\
    Gunning Fog               & 8.13  & 16.13 & 10.7\% \\
    SMOG                      & 8.84  & 14.73 & 10.5\% \\
    Dale--Chall               & 12.13 & 13.48 & 69.0\% \\
    Spache                    & 5.40  & 7.26  & 18.4\% \\
    Coleman--Liau             & 17.50 & 14.86 & 22.6\% \\
    ARI                       & 13.62 & 16.72 & 29.4\% \\
    Linsear Write             & 6.00  & 11.00 & 4.2\% \\
    LIX                       & 62.00 & 57.37 & 22.6\% \\
    RIX                       & 6.00  & 7.88  & 31.0\% \\
    FORCAST                   & 11.25 & 11.88 & 15.0\% \\
    Powers--Sumner--Kearl     & 6.31  & 7.48  & 8.1\% \\
    Words per sentence        & 12.00 & 22.75 & 55.6\% \\
    Syllables per word        & 1.67  & 1.74  & 6.2\% \\
    \midrule
    \multicolumn{4}{l}{\textit{Hohmann writing style suite (12 metrics)}} \\
    \midrule
    Sentence length (Hohmann) & --    & 31.42 & 87.2\% \\
    Parse depth               & --    & 7.00  & 58.9\% \\
    NP density                & --    & 0.23  & 5.3\% \\
    Noun chunks /100          & --    & 1.06  & 59.2\% \\
    Nouns /100                & --    & 29.44 & 50.9\% \\
    Verbs /100                & --    & 11.14 & 29.3\% \\
    Numbers /100              & --    & 9.81  & 100.0\% \\
    Sign posting /100         & --    & 0.00  & 50.0\% \\
    Hedging /100              & --    & 0.80  & 84.8\% \\
    Active narration          & --    & 1.00  & 100.0\% \\
    Passive rate              & --    & 0.08  & 26.2\% \\
    Type--token ratio         & --    & 0.64  & 8.5\% \\
    \midrule
    \multicolumn{4}{l}{\textit{Sensational language (10 categories, per 100 words)}} \\
    \midrule
    Importance                & --    & 0.00  & 72.3\% \\
    Novelty                   & --    & 0.00  & 73.9\% \\
    Rigor                     & --    & 0.00  & 69.9\% \\
    Scale                     & --    & 0.00  & 75.2\% \\
    Utility                   & --    & 0.00  & 67.2\% \\
    Quality                   & --    & 0.00  & 90.2\% \\
    Attitude                  & --    & 0.00  & 93.8\% \\
    Problem                   & --    & 0.00  & 99.3\% \\
    Additional emphasis       & --    & 0.00  & 85.7\% \\
    Total                     & --    & 0.00  & 19.9\% \\
    \midrule
    \multicolumn{4}{l}{\textit{Acronyms}} \\
    \midrule
    Acronym density /100      & 8.33  & 3.75  & 80.7\% \\
    Novel acronyms            & 0     & 0     & under P1 budget \\
    \bottomrule
  \end{tabular}
\end{table}

\end{document}